\documentclass[10pt,journal,compsoc]{IEEEtran}
\usepackage[nocompress]{cite}
\usepackage{lineno}
\usepackage{amsmath}
\usepackage{amssymb}
\usepackage{commath}
\usepackage{tabularx}
\usepackage{booktabs}
\usepackage{float}
\usepackage{algorithm}
\usepackage{algpseudocode}
\usepackage{graphicx}
\usepackage{subfigure}
\usepackage{color}
\usepackage{CJK}
\usepackage{cite}
\usepackage{multirow}
\usepackage[pagebackref=false,breaklinks=false,letterpaper=true,colorlinks,bookmarks=false]{hyperref}

\begin{document}
\title{A Performance Evaluation of Correspondence Grouping Methods for 3D Rigid Data Matching\IEEEcompsocitemizethanks{\IEEEcompsocthanksitem Jiaqi~Yang, Peng~Wang and Yanning~Zhang are with the School of Computer Science, Northwestern Polytechnical University and the National Engineering Laboratory for Integrated Aero-Space-Ground-Ocean Big Data Application Technology,  Xi'an 710129, China. E-mail: \{jqyang, peng.wang, ynzhang\}@nwpu.edu.cn. (\textit{Corresponding author: Peng Wang})
\IEEEcompsocthanksitem Ke Xian is with National Key Laboratory of Science and Technology on Multi-spectral Information Processing, School of Artificial Intelligence and Automation, Huazhong University of Science and Technology, Wuhan 430074, China. E-mail: kexian@hust.edu.cn.}}

\author{Jiaqi~Yang, Ke~Xian, Peng~Wang and Yanning~Zhang,~\IEEEmembership{Senior Member, IEEE}}

\markboth{Journal of \LaTeX\ Class Files,~Vol.~14, No.~8, August~2015}%
{Shell \MakeLowercase{\textit{et al.}}: Bare Demo of IEEEtran.cls for Computer Society Journals}
\IEEEtitleabstractindextext{%
\begin{abstract}
Seeking consistent point-to-point correspondences between 3D rigid data (point clouds, meshes, or depth maps) is a fundamental problem in 3D computer vision. While a number of correspondence selection methods have been proposed in recent years, their advantages and shortcomings remain unclear regarding different applications and perturbations. To fill this gap, this paper gives a comprehensive evaluation of nine state-of-the-art 3D correspondence grouping methods. A good correspondence grouping algorithm is expected to retrieve as many as inliers from initial feature matches, giving a rise in both precision and recall as well as facilitating accurate transformation estimation.   Toward this rule, we deploy experiments on three benchmarks with different application contexts including shape retrieval, 3D object recognition, and point cloud registration together with various perturbations such as noise, point density variation, clutter, occlusion, partial overlap, different scales of initial correspondences, and different combinations of keypoint detectors and descriptors. The rich variety of application scenarios and nuisances result in different spatial distributions and inlier ratios of initial feature correspondences, thus enabling a thorough evaluation.  Based on the outcomes, we give a summary of the traits, merits, and demerits of evaluated approaches and indicate some potential future research directions. 
\end{abstract}
\begin{IEEEkeywords}
Performance evaluation, correpondence grouping, 3D computer vision, 3D rigid data, shape matching.
\end{IEEEkeywords}}
\maketitle
\IEEEdisplaynontitleabstractindextext
\IEEEpeerreviewmaketitle

\IEEEraisesectionheading{\section{Introduction}\label{sec:intro}}
\IEEEPARstart{E}{stablishing} correct point-to-point matching relationship between rigid 3D shapes (e.g., point clouds, meshes, and depth maps), a.k.a. correspondence problem, is pivotal in 3D computer vision. It has been successfully applied to many areas such as point cloud registration~\cite{rusu2008aligning,yang2016fast}, object recognition~\cite{salti2014shot,guo2013rotational}, shape retrieval~\cite{boyer2011shrec}, and localization~\cite{tateno20162}. The key reason is that accurate  six-degree-of-freedom (6DoF) pose can be estimated with consistent 3D correspondences (matches), allowing data fusion, coordinate system normalization, and relative pose calculation. Moreover, when temporal information is available, we are able to obtain the movement information of a 3D object such as linear and angular velocities.

The initial correspondences between two shapes, namely a source shape and a target shape, are usually generated with three standard steps~\cite{mian2005automatic,guo20143d}. First, a set of sparse and distinctive keypoints are detected from the two shapes because the raw 3D data are always with great redundancy. Second, local geometric feature description is performed for each keypoint, specifically in the local surface patch formed by the radius neighbors of a keypoint, to encode the local shape and spatial information. Third, by matching local geometric feature descriptors using distance metrics such as Euclidean distance, we can obtain initial point-to-point correspondences.  However, this set may be contaminated by a large amount of false matches (outliers) due to the following reasons.

\begin{figure}[t]
	\centering
	\includegraphics[width=\linewidth]{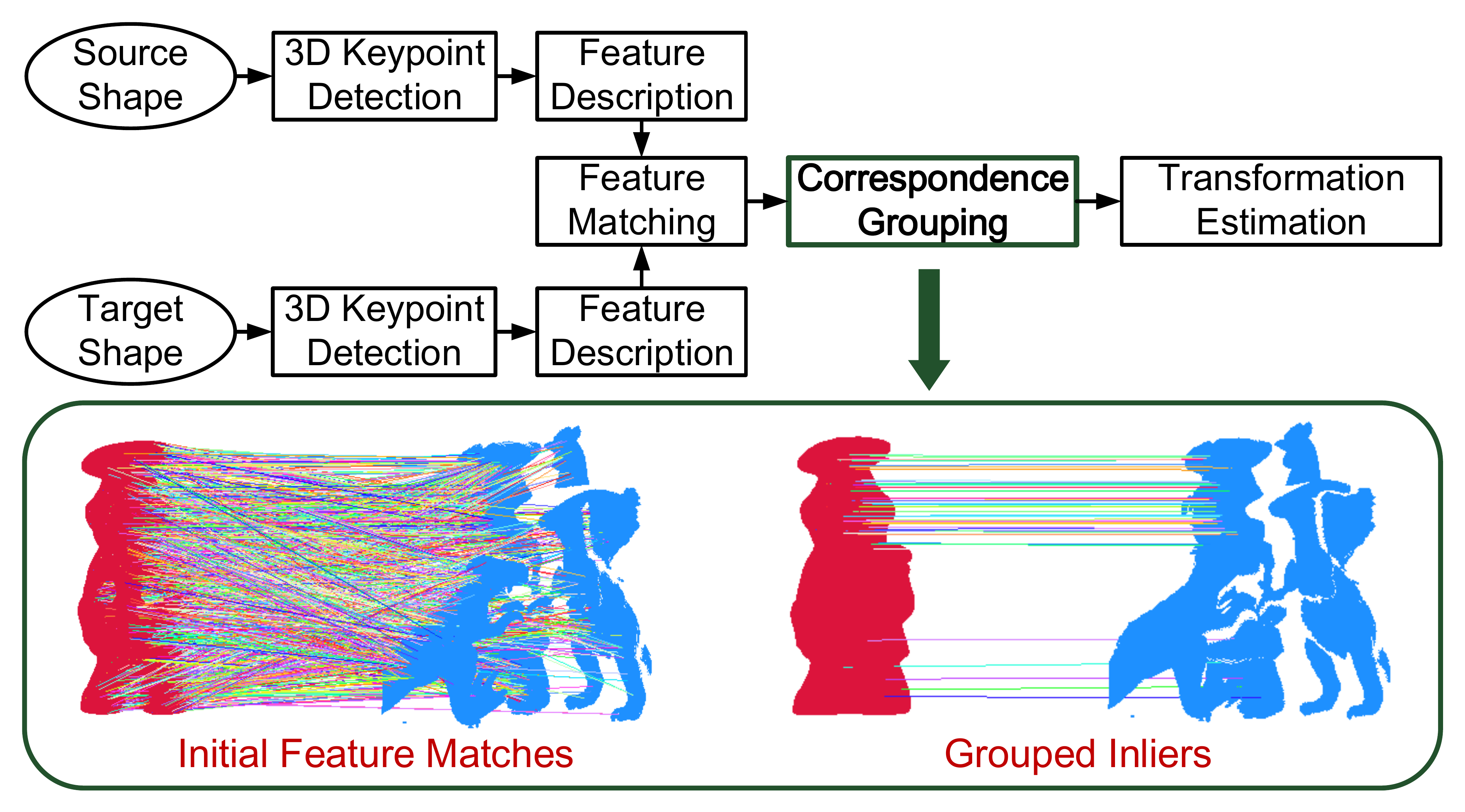}\\
	\caption{Illustration of local feature-based matching paradigm, where the objective of 3D correspondence grouping is searching inliers from an initial correspondence set with outliers between two shapes.}
	\label{fig:Mtd_illus}
\end{figure}
{\textbf{Data/scenario perturbations.}} {\textit{(i)}} Noise. Many current 3D sensors, especially for those recent low-cost Microsoft Kinect and Intel RealSense devices, are not able to capture clean point cloud data. Heavy noise will corrupt the geometric structure greatly.  {\textit{(ii)}} Data density variation. Distance changes between the sensor and the object/scene will lead to data resolution variation in raw data. This happens frequently for moving sensors or objects. {\textit{(iii)}} Clutter and occlusion. In 3D object recognition scenario, the object of interest is often localized in complex scenes with clutter and occlusion, causing data missing and huge amounts of outliers. {\textit{(iv)}} Partial overlap. In point cloud registration scenario, data scanned from different viewpoints usually suffer from limited overlaps. It naturally results in a number of correspondences that fall outside the overlapping region.

{\textbf{Limitations of detectors/descriptors.}} {\textit{(i)}} For existing 3D keypoint detectors, most of them still suffer from limited repeatability particularly in  point cloud registration and object recognition applications~\cite{tombari2013performance}. Truly corresponding points may miss each other because of keypoint localization errors. {\textit{(ii)}} For existing 3D local geometric descriptors, although a number of  descriptors have been proposed, evaluation studies~\cite{guo2016comprehensive,yang2017eval} still show that their performance needs improvement for data with severe occlusion and noise. In addition, local descriptors are sensitive to repetitive patterns.

The above challenges highlight the significance of correspondence grouping, making it a critical role in the  local feature-based matching paradigm (as shown Fig.~\ref{fig:Mtd_illus}) that is a prevalent approach for high-level vision tasks such as 3D registration and 3D object recognition. The goal of correspondence grouping is retrieving as many as inliers from raw feature correspondences, giving a rise in both precision and recall as well as facilitating accurate transformation estimation. In the 2D image domain, correspondence grouping is a long-standing issue with a number of solutions, e.g., parametric methods such as random sampling consensus (RANSAC)~\cite{fischler1981random} and  universal RANSAC (USAC)~\cite{raguram2013usac}; non-parametric methods such as grid-based motion statistics (GMS)~\cite{bian2017gms} locality preserving matching (LPM)~\cite{ma2018locality}. By directly changing the hypothesis model for parametric methods or leveraging 3D geometric constraints for non-parametric methods, many of these methods can be generalized to the 3D domain, e.g., RANSAC and spectral technique (ST)~\cite{leordeanu2005spectral}. However, 3D data possess some unique peculiarities to images, e.g.,  surface normals, local reference frame (LRF), and  rigidity. Through leveraging these properties, a number of 3D-targeted correspondence grouping methods have been proposed~\cite{chen20073d,mian2010repeatability,tombari2010object,rodola2013scale,buch2014search,yang2019ranking,bustos2018guaranteed}. Based on the number of candidates considered for grouping, we categorize them as \textit{individual}-based and \textit{group}-based. Individual-based methods judge the correctness of a correspondence independently, while group-based methods try to find the cluster formed by inliers. With the wealth of a wide range of 3D correspondence grouping
techniques, yet, the advantages and shortcomings of them remain unclear because the efficacy of each method is usually tested on datasets addressing a particular application with limited variety in nuisance types and insufficient comparisons, making it confusing for developers to select a proper method for a specific application.

Motivated by these considerations, this paper presents a comprehensive evaluation of nine popular 3D correspondence grouping algorithms, i.e., similarity score (SS), nearest neighbor similarity ratio (NNSR)~\cite{lowe2004distinctive}, ST~\cite{leordeanu2005spectral}, RANSAC~\cite{fischler1981random}, geometric consistency (GC)~\cite{chen20073d}, 3D Hough voting~\cite{tombari2010object}, game theory matching (GTM)~\cite{rodola2013scale}, search of inliers (SI)~\cite{buch2014search}, and consistency voting (CV)~\cite{yang2019ranking}. SS and NNSR are considered as baselines because one can directly group matches based on the local descriptor's distinctiveness. As the input for a correspondence grouping method, the initial correspondence set varies in terms of \textit{inlier ratios} and \textit{spatial distributions}. To promise a rich variety of the two terms, we consider three application contexts including shape retrieval, 3D object recognition, and point cloud registration with different nuisances including noise, point density variation, clutter, occlusion, partial overlap, different scales of initial correpsondences and different combinations of keypoint detectors and descriptors. All considered nuisances have been quantized for a detailed comparison. The efficiency of tested methods respecting correspondence sets with different scales has also been tested. This paper extends the conference version~\cite{yang2017performance} from five aspects. {\textit{(i)}} More evaluated methods. This paper additionally considers two methods, i.e., the well-known GTM~\cite{rodola2013scale} and recently proposed CV~\cite{yang2019ranking}, to achieve a more extensive comparison. {\textit{(ii)}} More metrics. In addition to precision and recall, we also consider the F-score and rigid data  alignment measure to achieve aggregated and task-level evaluation. {\textit{(iii)}} More analysis on experimental data. We present detailed statistics of the number of inliers and inlier ratios for each experimental data setting. It enables an insightful illustration of the effect caused by each considered nuisance on the quality of initial matches. {\textit{(iv)}} More experiments.  Rather than using a fixed detector-descriptor combination~\cite{yang2017performance}, this paper considers six combinations of 3D keypoint detectors and descriptors to address the concern of the effect when using different detector-descriptor combos. {\textit{(v)}} A review of the existing literature. Existing evaluation works related to this evaluation have been comprehensively surveyed and discussed. To summarize, this paper has three main contributions.
\begin{itemize}
	\item An abstraction of  nine state-of-the-art 3D correspondence grouping methods into a set of core stages, which helps to highlight the peculiarities of each approach and their differences.
	\item A comprehensive performance evaluation of several popular 3D correspondence grouping methods. This evaluation covers the major concerns for such topic,  performance under a large variety of application contexts and perturbations (e.g., noise, point density variation, clutter, occlusion, partial overlap, different numbers of initial correspondences, and different detector-descriptor combinations) and computational efficiency.
	\item A summary and discussion of evaluated methods in terms of their traits, merits, and demerits based on the experimental outcomes. This provides useful application guidance for developers and  potential future research directions for scholars to overcome existing issues in this research field.
\end{itemize}

The remainder of this paper is organized as follows. Sect.~\ref{sec:related} reviews related evaluations in the area of local geometric feature matching. Sect.~\ref{sec:mtd} gives a taxonomy and review of nine state-of-the-art methods by identifying the core computational steps of each method. Sect.~\ref{sec:eval_mtd} presents the evaluation methodology with descriptions on datasets, metrics, considered terms for evaluation, and implementation details of each method. Sect.~\ref{sec:result} presents the evaluation results and relevant explanations. Sect.~\ref{sec:sum} gives a summary and discussion of for each method based on the evaluation results. The conclusions are finally drawn in Sect.~\ref{sec:conc}.
\section{Related Work}\label{sec:related}
This section reviews existing performance evaluation works relate to local geometric feature-based matching.

For 3D keypoint detection, Tombari et al.~\cite{tombari2013performance} conducted an evaluation of several fixed-scale and adaptive 3D keypoint detectors in terms of repeatability, distinctiveness, and computational efficiency. The evaluation datasets are characterized by interferences such as noise, clutter, occlusions, and partial overlap.  For 3D local feature descriptors, Sukno et al.~\cite{sukno2012comparing} presented a comparison of local geometric descriptors for craniofacial landmarks annotated on 144 point cloud scans for clinical research. Kim and Hilton~\cite{kim2013evaluation} tested four 3D local feature descriptors in the context of multi-modal data registration, showing that the robustness of current descriptors to data modality changes is still limited. Guo et al.~\cite{guo2016comprehensive} quantitatively assessed ten popular 3D local geometric feature descriptors on datasets addressing shape retrieval, point cloud registration, and 3D object recognition scenarios. The evaluated terms including descriptiveness, compactness, efficiency, and robustness to a number of common nuisances.  Buch et al.~\cite{buch2016local} performed an evaluation of the results when fusing various local geometric features for the task of 3D object recognition, showing that the feature matching, pose estimation, and object recognition performance can be effectively boosted with proper combinations of geometric features. Yang et al.~\cite{yang2017eval} evaluated the effect of different characterizations of the local spatial information on the distinctiveness, compactness, and robustness of local descriptors.  Because many of existing local geometric features rely on a local reference frame (LRF) on one hand to achieve rotation invariance and on the other to fully characterize the local spatial information, there are also some evaluations on LRFs. Specifically,  Petrelli and Stefano~\cite{petrelli2011repeatability} presented a study on the effect of LRF errors on the distinctiveness of local descriptors and gave an evaluation of seven LRFs in the context of partial shape matching. Later, they proposed a new metric to more precisely evaluate the repeatability of LRFs and tested two additional LFRs with experiments on a vast corpus of data. Yang et al.~\cite{yang2018toward} classified existing LRFs as covariance analysis-based and point spatial distribution-based and tested a total of eight LRFs on six datasets with different application scenarios and nuisances. They also assessed the descriptor matching performance when using different LRFs and feature representation combinations. 
In addition to independently evaluating keypoint detectors and descriptors, Bronstein et al.~\cite{bronstein2010shrec} and Boyer et al.~\cite{boyer2011shrec} tested the performance of shape feature detectors and descriptors under a wide range of transformations for the task of shape retrieval. Salti et al.~\cite{salti2012affinity} compared many possible combinations between state-of-the-art 3D detectors and descriptors on datasets with different modalities and contexts. H{\"a}nsch et al.~\cite{hansch2014comparison} evaluated the effectiveness of different combinations of 3D keypoint detectors and descriptors for point cloud fusion/reconstruction. The considered datasets including LiDAR and Kinect point clouds. For geometric feature matching, Rusu et al.~\cite{rusu2008persistent} analyzed  six distance metrics for the matching of point feature histograms in point clouds when applied to point-to-point correspondences searching and point classification. These distance metrics were also evaluated in~\cite{yang2017RCS_jrnl} and~\cite{buch2018local} for the matching of rotational contour signatures (RCS) and local point pair feature histograms (PPF Hist), respectively.

Regarding the point-to-point correspondence problem,  Raguram~\cite{raguram2008comparative} evaluated the RANSAC algorithm and several of its variants for robust  estimation from image feature correspondences with outliers. Bian et al.~\cite{bian2018matchbench} gave a comparative evaluation of several 2D feature matchers and proposed a uniform feature matching benchmark. These matchers were evaluated from different aspects including matching ability, correspondence sufficiency, and efficiency. {\textit{However, to the best of our knowledge, no prior works have been carried out for the evaluation of  correspondence grouping approaches in the 3D domain (except the preliminary version of this paper~\cite{yang2017performance}) and they mainly concentrate on studying feature detectors and descriptors.}} A comprehensive evaluation of 3D correspondence grouping methods is necessary and valuable because it will advance the development of this field and provide complementary information to existing evaluations about local geometric feature detectors and descriptors.
\section{Considered  Methods}\label{sec:mtd}
Nine 3D correspondence grouping algorithms are considered in our evaluation mainly die to their popularity and state-of-the-art performance. They are either individual-based or group-based. Before recapping these methods, we first describe some notations for better readability. 

The correspondence grouping problem can be formulated as follows. Given a source shape  $\cal{S}$  and a target shape $\cal{S'}$,  an initial correspondence set $\cal C$ is generated by matching the feature sets $\cal F$ and $\cal F'$  computed for the keypoints on $\cal{S}$ and $\cal{S'}$, respectively. The goal is to seek an inlier set ${\cal C}_{inlier}$ from $\cal C$ that identifies the correct matching relationship between $\cal{S}$ and $\cal{S'}$. A component in $\cal C$ can be parameterized by $c = \{ {\bf p},{\bf p}', s_{\cal F}({\bf f},{\bf f}')\} $, where ${\bf p} \in {\cal S}$, ${\bf p}' \in {\cal S'}$, ${\bf f} \in {\cal F}$, ${\bf f}' \in {\cal F'}$, and $s_{\cal F}({\bf f},{\bf f}')$ is the feature similarity score assigned to $c$. With these notations, we describe the key ideas and  computation steps of each method in the following.
\subsection{Individual-based}
Individual-based methods intend to first assign a score for each correspondence using feature similarity or geometric cues, and then  group correspondences independently based on the scores. The score can be computed either using a single correspondence or taking local/global context information into consideration.
\\
\\
\noindent\textbf{Similarity Score.} The initial correspondence set can be naively split based on the similarity score $s_{\cal F}({\bf f},{\bf f}')$ of the two descriptors extracted from $( {\bf p},{\bf p}')$~\cite{mian2006three,yang2016fast}. The assumption is that correspondences agreed with more similar descriptors are more likely to be correct. Although a number of distinctive 3D local features~\cite{tombari2010unique,guo2013rotational} have been proposed, common nuisances such as noise, data resolution variations, and repetitive patterns could easily cause false judgments. This method, dubbed as SS, is served as a baseline in our evaluation that judges a correspondence as correct if:
\begin{equation}
{1-\| {{\bf f}_{norm} - {\bf f}'_{norm}} \|_{{L_2}}} \ge {t_{ss}},
\end{equation}
where ${\bf f}_{norm}$ is the normalized feature of ${\bf f}$ and $t_{ss} \in [0,1]$.  We use the $L_2$ distance to calculate $s_{\cal F}({\bf f},{\bf f}')$ in this paper.
\\
\\
\noindent\textbf{Nearest Neighbor Similarity Ratio~\cite{lowe2004distinctive}.} Lowe's ratio rule~\cite{lowe2004distinctive} is regarded as the other baseline in this evaluation. It penalizes correspondences by the ratio of the nearest and the second nearest distance in feature space. This rule enables \textit{distinctive} regions to achieve high ranking scores. Analogous to SS's thresholding strategy, NNSR accepts a correspondence as inlier if:
\begin{equation}\label{eq:ratio}
1 - \frac{{{{\left\| {{\bf f} - {\bf f}_{nn1}^{'}} \right\|}_{{L_2}}}}}{{{{\left\| {{\bf f} - {\bf f}_{nn2}^{'}} \right\|}_{{L_2}}}}} \ge {t_{nnsr}},
\end{equation}
where ${\bf f}_{nn1}^{'}$ and ${\bf f}_{nn2}^{'}$ respectively represent the most and the second most similar features to $f$, and $t_{nnsr} \in [0,1]$.
\\
\\
\noindent\textbf{Search of Inliers~\cite{buch2014search}.} The search of inliers (SI)~\cite{buch2014search} method concentrates on the  problem of 3D correspondence selection for 3D rigid data matching. SI follows a voting paradigm and designs both spatially local and global geometric constraints to determine whether a vote should be casted. This method includes three main steps, i.e, initialization, local voting, and global voting.

During initialization, a portion of the initial correspondence set $\cal C$, i.e., ${\cal C}_{Ratio}$,  is selected with the Lowe's ratio rule (c.f. Eq.~\ref{eq:ratio}). At the local voting stage, local voters for $c$ are defined as the intersection of ${\cal C}_{Ratio}$ and the nearest $\kappa$  neighbors of $c$. The components in ${\cal C}_L(c)$ that satisfy the rigidity constraint (c.f. Eq.~\ref{eq:rigid}) are defined as the positive local votes $\Upsilon_L(c)$:
\begin{equation}
\Upsilon_L(c)=\{c_L \in {\cal C}_L(c) : r(c,c_L)>\varsigma\},
\end{equation}
where $\varsigma$ is a free parameter at the  local voting stage. The local score of $c$ is defined as $s_L(c)=\frac{|\Upsilon_L(c)|}{|{\cal C}_L(c)|}$.

At the global voting stage, the global voters ${\cal C}_G$ are selected as the former $\kappa$ correspondences ranked in a decreasing order according to Lowe's ratio scores. To judge the affinity between two correspondences, $c_i$ and $c_j$ will take the following test:
\begin{equation}
v_G(c_i,c_j)=d({\bf T}(c_i) \cdot {\bf p}_j, {\bf p}'_j),
\end{equation}
where ${\bf T}(c) $ is ${\bf{R}(p')}^{-1} \cdot {\bf{R}(p)}$ with ${\bf{R}(p)}$ being the LRF of ${\bf p}$. The global voters are then localized by applying both local and global constraints:
\begin{equation}
\Upsilon_G(c)=\{c_G \in {\cal C}_G : r(c,c_G)>\varsigma  \wedge v_G(c,c_G) < \delta \},
\end{equation}
where $\delta$ is a Euclidean distance tolerance. The eventual vote score for $c$ is defined as:
\begin{equation}
s(c)=\frac{|\Upsilon_L(c)|+|\Upsilon_G(c)|}{|{\cal C}_L(c)|+|{\cal C}_G(c)|}.
\end{equation}

The correspondences with higher values than the threshold determined based on Otsu's adaptive method~\cite{otsu1975threshold} are SI-judged inliers.
\\
\\
\noindent\textbf{Consistency Voting~\cite{yang2019ranking}.} Based on the assumption that only inliers are compatible with each other, Yang et al.~\cite{yang2019ranking} proposed consistency voting (CV) to check the agreement of a query correspondence with a voting set composed by distinctive correspondences. There are three main computational steps.

First, the initial correspondence set is ordered according to NNSR~\cite{lowe2004distinctive} scores and top-$k$ candidates from the re-ordered set are served as the voting set. Second, a compatibility measure for a correspondence pair $(c_i,c_j)$ that incorporates both rigidity and LRF affinity constraints is introduced. The rigidity term is defined as:
\begin{equation}
\begin{aligned}
r({c_i},{c_j}) =\left|\|{{\bf p}_i}-{{\bf p}_j}\|_{L_2} - \|{{\bf p}'_i}-{{\bf p}'_j}\|_{L_2}\right|.
\end{aligned}
\end{equation}
The LRF affinity term is defined as:
\begin{equation}
\begin{aligned}
L(c_i,c_j)=\left|\epsilon_{lrf}({\bf p}_i,{\bf p}_j)-\epsilon_{lrf}({\bf p}'_i,{\bf p}'_j) \right|,
\end{aligned}
\end{equation}
where
\begin{equation}
\begin{aligned}
\epsilon_{lrf}({\bf p}_i,{\bf p}_j){\rm{ = }}{\mathop{\rm acos}\nolimits} \left( {\frac{{{{\rm{trace}}{\left({{{\bf L}_{{\bf p}_i}}{{{\bf L}_{{\bf p}_j}}}^{-1}}\right) - 1}}}}{2}} \right)\frac{{180}}{\pi },
\end{aligned}
\end{equation}
where ${{\bf L}_{{\bf p}_i}}$ represents the LRF of ${\bf p}_i$. By combining both constraints, the compatibility measure is defined as:
\begin{equation}
\begin{array}{l}
{\cal D}({c_i},{c_j}) ={\rm exp}(-\frac{r({c_i},{c_j})^2}{\delta_r^2}-\frac{L(c_i,c_j)^2}{\delta_L^2}),
\end{array}
\end{equation}
where $\delta_r$ and $\delta_L$  represent the rigidity and LRF affinity parameters, respectively.

Third, the final voting score of a correspondence $c$ is defined as the aggregation of all compatibility scores of $c$ and $c_i \in {\cal C}_v$:
\begin{equation}\label{eq:cv_score}
s(c) = \sum\nolimits_{{c_i} \in {{\cal C}_v}} {{\cal D}(c,{c_i})} .
\end{equation}
Correspondences with higher voting scores than a threshold $t_{cv}$ are served as inliers.
\subsection{Group-based}
Group-based methods assume that inliers form a cluster in a particular domain and select correct correspondences in an one-shot manner. The finding of inlier cluster usually relies on generating reasonable hypotheses, seeking cluster center or candidates that pairwisely agree with each other.
\\
\\
\noindent\textbf{Random Sampling Consensus~\cite{fischler1981random}.} RANSAC~\cite{fischler1981random} iteratively performs hypothesis-verification and evaluates the correctness of current samples based on the number of identified inliers. It has been broadly employed in both 2D~\cite{brown2007automatic} and 3D domains~\cite{rusu2008aligning}. Despite its variants~\cite{torr2000mlesac,chum2005matching}, we focus on evaluating the original RANSAC method.

It repeatedly performs the following operations $N_{ransac}$ times.
At each iteration, the method first randomly samples three candidates from $\cal C$.  Second, the sampled correspondences are used to generate a hypothesis, i.e., a transformation ${\bf{T}}_i \in SE(3)$ for 3D rigid data alignment. To judge the correctness of ${\bf{T}}_i$, the source keypoints in $\cal C$ (the intersection of $\cal S$ and $\cal C$) will be transformed using ${\bf{T}}_i$. The confidence of ${\bf{T}}_i$ is positively correlated to  the number of transformed source keypoints whose Euclidean distances to their corresponding points in $\cal S'$ are smaller than a threshold $d_{ransac}$. Finally, the transformation yielding to the maximum inlier count is taken as the optimal ${\bf{T}}^*$, and correspondences in $\cal C$ coherent with ${\bf{T}}^*$ are grouped as inliers.
\\
\\
\noindent\textbf{Spectral Technique~\cite{leordeanu2005spectral}.} Spectral methods are widely used for searching the main cluster of a graph~\cite{shi2000normalized,mahamud2003segmentation}. Leordeanu and Hebert~\cite{leordeanu2005spectral} used a spectral technique (ST) to group correspondences based on the observation that inliers in $\cal C$ should form a consistent cluster. The essential idea is finding the level of \textit{association} of each correspondence with the main cluster exits in the initial correspondence set $\cal C$. The basic calculation procedure can be abstracted as follows.

First, an adjacent non-negative matrix $\bf M$ is computed for $\cal C$ in which each component is the pairwise affinity score of two correspondences.  Second, the principle eigenvector of $\bf M$ is calculated as $\bf v$ and the location of the maximum value of $\bf v$, e.g., ${\bf v}_i$, indicates $c_i$ being the cluster center. Third, remove from $\cal C$ all potential components in conflict with $c_i$. By repeating the previous two steps until ${\bf v}_i =0$ or $|{\cal C}|=0$, the selected candidates from the second step thus consist the final inlier cluster.

ST is general for both 2D and 3D correspondence problems, depending on the definition of the pairwise affinity term. Here, we use the popular rigidity constrain~\cite{johnson1998surface,buch2014search} in the 3D domain as the pairwise term of $c_i$ and $c_j$, which is defined as:
\begin{equation}\label{eq:rigid}
r ({c_i},{c_j}) = \min ( {\frac{{{{\| {{{\bf p}_i},{{\bf p}_j}} \|}_{{L_2}}}}}{{{{\| {{\bf p}'_i,{\bf p}'_j} \|}_{{L_2}}}}},\frac{{{{\| {{\bf p}'_i,{\bf p}'_j} \|}_{{L_2}}}}}{{{{\| {{{\bf p}_i},{{\bf p}_j}} \|}_{{L_2}}}}}} ).
\end{equation}
By thresholding on $r ({c_i},{c_j})$ using $t_{st}$, one can conclude whether $c_i$ and $c_j$ are compatible or not.
\\
\\
\noindent\textbf{Geometric Consistency~\cite{chen20073d}.} The GC method~\cite{chen20073d} is independent from the feature space and applies constraints relating to the compatibility of spatial locations of corresponding points. The compatibility score for two given correspondences $c_i$ and $c_j$ is defined as:
\begin{equation}\label{eq:gc}
d({c_i},{c_j}) = | {d({{\bf p}_i},{{\bf p}_j}) - d({{\bf p}'_i},{{\bf p}'_j})} | < {t_{gc}},
\end{equation}
where $d({{\bf p}_i},{{\bf p}_j})={\|{\bf p}_i - {\bf p}_j \|}_{L_2}$ and $t_{gc}$ is a threshold to judge if $c_i$ and $c_j$ satisfy the geometric constraint or not.

With the above rule, GC then associates a \textit{consistent} cluster to each correspondence. Particularly, for a correspondence $c$, its compatibility scores with all other correspondences in $\cal C$ are computed based on Eq.~\ref{eq:gc}. All the correspondences with confirmed compatibility scores therefore form a cluster for $c$, and the size of the cluster decides how the current cluster is likely to be the inlier cluster. By repeating the procedure for all correspondences, the largest cluster is served as the grouped inlier set.
\\
\\
\noindent\textbf{3D Hough Voting~\cite{tombari2010object}.} The Hough transform~\cite{vc1962method} is a popular computer vision technique initially proposed to detect lines in images. Tombari and Stefano~\cite{tombari2010object} extended it for 3D object recognition and named it as 3D Hough voting (3DHV) . This method is also employed to group correspondences for partial shape matching~\cite{petrelli2015pairwise}. In 3DHV, each correspondence casts a vote in a 3D Hough space based on the following steps.

For the $i$th correspondence in $\cal C$ denoted by $c_i = \{ {\bf p}_i,{\bf p}'_i\} $, the vector between ${\bf p}_i \in {\mathbb{R}^3} $ and the  centroid ${\bf c}_{\cal S} \in {\mathbb{R}^3}$ of the source shape $\cal S$ is firstly computed as:
\begin{equation}
{\bf{V}}_{i,G}^{\cal S} = {{\bf c}_{\cal S}} - {{\bf p}_i},
\end{equation}
which is then transformed in the coordinates given by the local reference frame (LRF) of ${\bf p}_i$ as:
\begin{equation}
{\bf{V}}_{i,L}^{\cal S} = {\bf{R}}_i^{\cal S}  {\bf{V}}_{i,G}^{\cal S},
\end{equation}
where ${\bf{R}}_i^{\cal S}$ is the rotation matrix and each row of ${\bf{R}}_i^{\cal S}$ is a unit vector of the LRF at ${\bf p}_i$. LRF is an independent coordinate system constructed in the local surface around a keypoint for the purpose of making the underlying feature representation rotation invariant and leveraging full spatial information. This step endows the vector of ${\bf p}_i$ with invariance to rigid transformations. Analogously, we can obtain a vector ${\bf{V}}_{i,L}^{\cal S'}$ for ${\bf p}'_i$. If ${\bf p}_i$ and ${\bf p}'_i$ are correctly corresponded, ${\bf{V}}_{i,L}^{\cal S'}$ should coincide with ${\bf{V}}_{i,L}^{\cal S}$. Based on this assumption, the vector ${\bf{V}}_{i,L}^{\cal S'}$ is finally transformed in the global coordinate of $\cal S'$ as:
\begin{equation}
{\bf{V}}_{i,G}^{\cal S'} = {\bf{R}}_i^{\cal S'}  {\bf{V}}_{i,L}^{\cal S'}+{\bf p}'_i.
\end{equation}
With above transformations, the feature ${\bf f}'_i$ could  cast a vote in a 3D Hough space by means of a vector ${\bf{V}}_{i,G}^{\cal S'}$. The peak in the Hough space indicates center of the cluster constituted by inliers. 
\\
\\
\noindent\textbf{Game Theory Matching~\cite{rodola2013scale}.} Rodol{\` a} et al.~\cite{rodola2013scale} interpreted the correspondence grouping problem as a non-cooperative game and proposed game theory matching (GTM) that  selects a small group of highly coherent correspondences. In the GTM framework, candidates in the initial correspondence set $\cal C$  are treated as available strategies. Pairs of players play a symmetric game and will adapt their behavior to prefer strategies that receive larger payoffs. Let ${\bf x}=(x_1,\cdots,x_{|\cal C|})^T$ represents the amount of population that plays each strategy $c_i$ at a given time. At the beginning, the initial population is set around the barycenter to be fair for each strategy. Then, the population will dynamically updates ($N_{gtm}$ repetitions in our evaluation) with an evolutionary process by applying the following  replicator dynamics equation:
\begin{equation}
{{\bf{x}}_i}(t + 1) = {{\bf{x}}_i}(t)\frac{{{{({\mathit \Pi} {\bf{x}}(t))}_i}}}{{{\bf{x}}{{(t)}^T}{\mathit \Pi}{\bf{x}}(t)}},
\end{equation}
where ${\mathit \Pi}$ is the payoff matrix that assigns the payoff between strategies $c_i$ and $c_j$ to row $i$ and column $j$. Such dynamics will converge to a Nash equilibrium~\cite{weibull1997evolutionary}. The payoff between two correspondences $c_i$ and $c_j$ is measured by the compatibility between them. GTM also employs the rigidity-based compatibility metric presented in Eq.~\ref{eq:rigid}. Therefore, the payoff matrix is defined as:
\begin{equation}
{\mathit \Pi} = \left\{ {\begin{array}{*{20}{c}}
	{r({c_i},{c_j}),}&{{\rm{if }}~{c_i} \ne {c_j}}\\
	{0,}&{{\rm{otherwise}}{\rm{.}}}
	\end{array}} \right.
\end{equation}

After the evolutionary process, correspondences (strategies) with larger populations, i.e., the population playing a strategy is greater than a threshold $t_{gtm}$, are served as inliers.
\section{Evaluation Methodology}\label{sec:eval_mtd}
This section introduces the experimental datasets, performance metrics, considered nuisances, and implementation details of tested methods. We intend to generate inputs for correspondence grouping with various inlier ratios and spatial distributions, achieved by matching rigid data from different application scenarios or injected with different nuisances.

\subsection{Datasets}
\begin{figure}[b]
	\centering
	\includegraphics[width=1.0\linewidth]{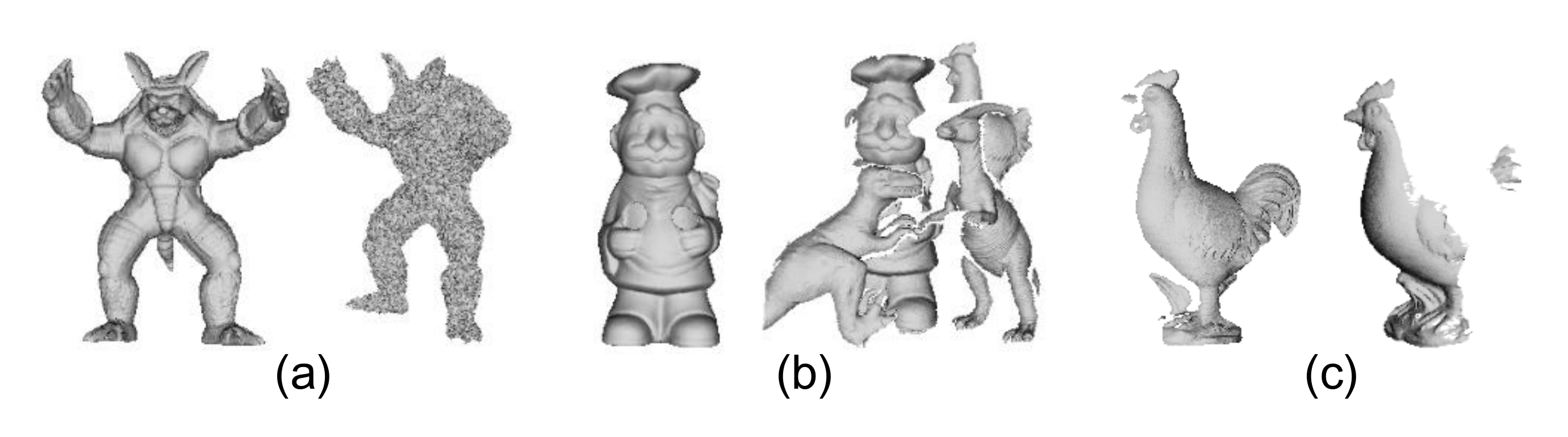}\\
	\caption{Sample views (visualized in mesh representation) from (a) B3R~\cite{tombari2013performance}, (b) U3OR~\cite{mian2006three,mian2010repeatability}, and (c) U3M~\cite{mian2006novel} datasets.}
	\label{fig:dataset}
\end{figure}
\noindent\textbf{B3R~\cite{tombari2013performance}.} The Bologna 3D Retrieval (B3R) dataset addresses the scenario of retrieving 3D rigid complete shapes. The original B3R dataset has three types of data group corrupted by  Gaussian noise with 0.1 \textit{pr}, 0.3 \textit{pr}, and 0.5 \textit{pr} standard deviations. Here and hereinafter,  \textit{pr} denotes the point cloud resolution, i.e., the average shortest distance among neighboring points in the point cloud. For each data group, 8 models and 18 scenes are included where scenes are rotated copies of models. The models are all taken from the Stanford Repository~\cite{curless1996volumetric}. The original B3R dataset therefore provides 54 matching pairs. In our evaluation, we will augment the B3R dataset by considering 9 levels of Gaussian noise and down-sampling the scenes to generate 9 data decimation levels. The augmented B3R dataset finally contains 324 scenes with quantized levels of noise and point density variation.
\\
\\
\noindent\textbf{U3OR~\cite{mian2006three,mian2010repeatability}.} The UWA 3D object recognition (U3OR) dataset is a popular dataset in 3D computer vision that addresses model-based 3D object recognition with clutter and occlusion. It has 5 models and 50 scenes. The scenes are generated by first randomly placing 4 or 5 models together and then scanning them from a particular view, resulting in approximately 65\%-95\%  clutter and 60\%-90\% occlusion. A total of 188 valid matching pairs are available in this dataset.
\\
\\
\noindent\textbf{U3M~\cite{mian2006novel}.} The  UWA 3D modeling (U3M) addresses point cloud (2.5D views) registration scenarios. It consists of 22, 16, 16 and 21 2.5D views respectively captured from the \textit{Chef}, \textit{Chicken}, \textit{T-rex}, and \textit{Parasaurolophus} models. Since the ground truth transformations are not available for this dataset, we conduct manual alignment for each data pair and then use ICP~\cite{besl1992method} for refinement to obtain the ground truth. Eventually, 425 valid data pairs that have at least 30\% overlap are available for evaluation.
\begin{figure}[t]
	\centering
	\includegraphics[width=1.0\linewidth]{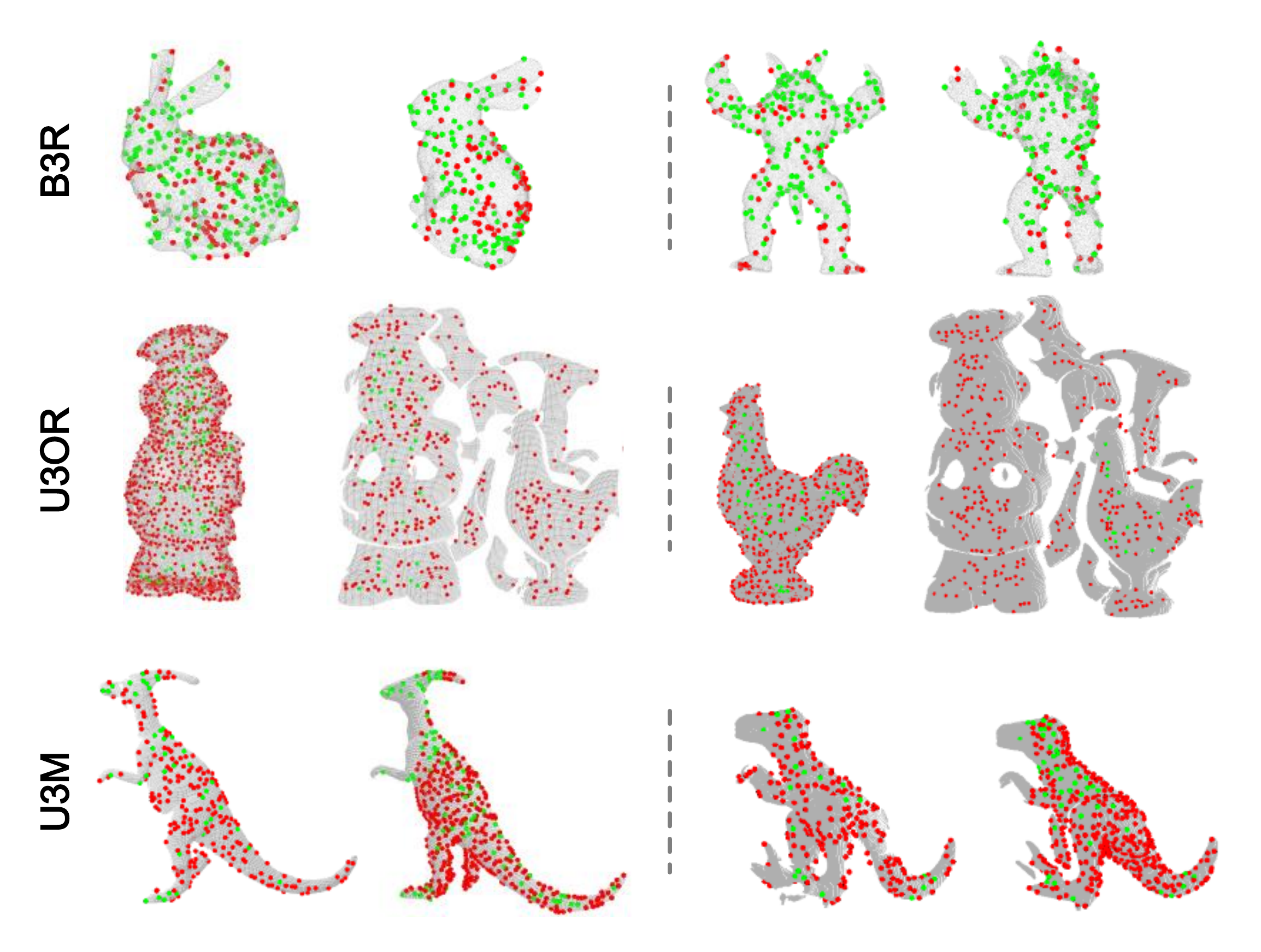}\\
	\caption{Spatial distributions of corresponding keypoints (detector: H3D~\cite{sipiran2011harris}; descriptor: SHOT~\cite{tombari2010unique}) on two example matching pairs for each dataset. Green and red dots represent keypoints with  correct and false correspondences, respectively.}
	\label{fig:corre_distri}
\end{figure}

The experimental datasets (Fig.~\ref{fig:dataset}) have different application scenarios, particularly covering ``full to full'' (B3R), ``full to partial'' (U3OR), and ``partial to partial'' (U3M) matching cases. It makes the initial correspondences generated by matching keypoint descriptors on these datasets distribute spatially different and contain various numbers of inliers. The variety of initial correspondences can be further enriched by changing the sparsity of keypoints or trying different \textit{detector-descriptor} combinations, enabling a thorough evaluation for correspondence grouping.
\subsection{Criteria}\label{sec:metric}
We judge the quality of a correspondence grouping method from two perspectives. One is the quality of the reduced correspondence set and the other is whether successful rigid registration can be accomplished using the reduced set. 

Regarding the former one, we use \textit{precision}, \textit{recall}, and \textit{F-score} for assessment. Let ${\bf{T}}_{GT} =\{ {\bf{R}}_{GT},  {\bf{t}}_{GT}  \}$ denote the ground truth transformation between $\cal S$ and $\cal S'$, where ${\bf{T}}_{GT}\in SE(3)$, ${\bf{R}}_{GT}\in SO(3)$, and ${\bf{t}}_{GT}\in {\mathbb{R}^3}$. A correspondence $c=(p,p')$ is accepted as correct only if:
\begin{equation}\label{eq:judge}
{\|  p\cdot {\bf{R}}_{GT} + {\bf{t}}_{GT} - p'  \|}_{L_2} \le \epsilon,
\end{equation}
where $\epsilon$ is a distance threshold. Let ${\cal C}_{group}$, ${\cal C}_{group}^{inlier}$, and ${\cal C}_{GT}$ respectively denote the grouped correspondence set by a tested method, inliers in the grouped set, and the ground truth inlier set in the initial set $\cal C$, \textit{precision} and \textit{recall} are defined as:
\begin{equation}
{\rm{Precision}} =\frac{|{\cal C}_{group}^{inlier}|}{|{\cal C}_{group}|},
\end{equation}
\begin{equation}
{\rm{Recall}} =\frac{|{\cal C}_{group}^{inlier}|}{|{\cal C}_{GT}|},
\end{equation}
and \textit{F-score} is given as $\frac{2\rm{Precision}\cdot\rm{Recall}}{\rm{Precision}+\rm{Recall}}$.

Regarding the latter one, as previously illustrated in Fig.~\ref{fig:Mtd_illus}, the following procedure after correspondence grouping is performing transformation estimation for data alignment and fusion. The de-facto estimator for 3D rigid data matching is arguable RANSAC~\cite{guo20143d}. RANSAC prefers correspondences with high precision because 3 correct correspondences are sufficient to achieve successful registration~\cite{johnson1999using}. Therefore, we judge a registration as successful if the precision of the grouped correspondence set is greater than a threshold $\tau_{reg}$. We experimentally observe that RANSAC manages to quickly estimate a reasonable transformation when the correspondence set have at least 10\% inliers. Thus, we set $\tau_{reg}$ to 0.1.
\subsection{Challenges}\label{sec:challenges}

\begin{figure}[t]
	\centering
	\includegraphics[width=1.0\linewidth]{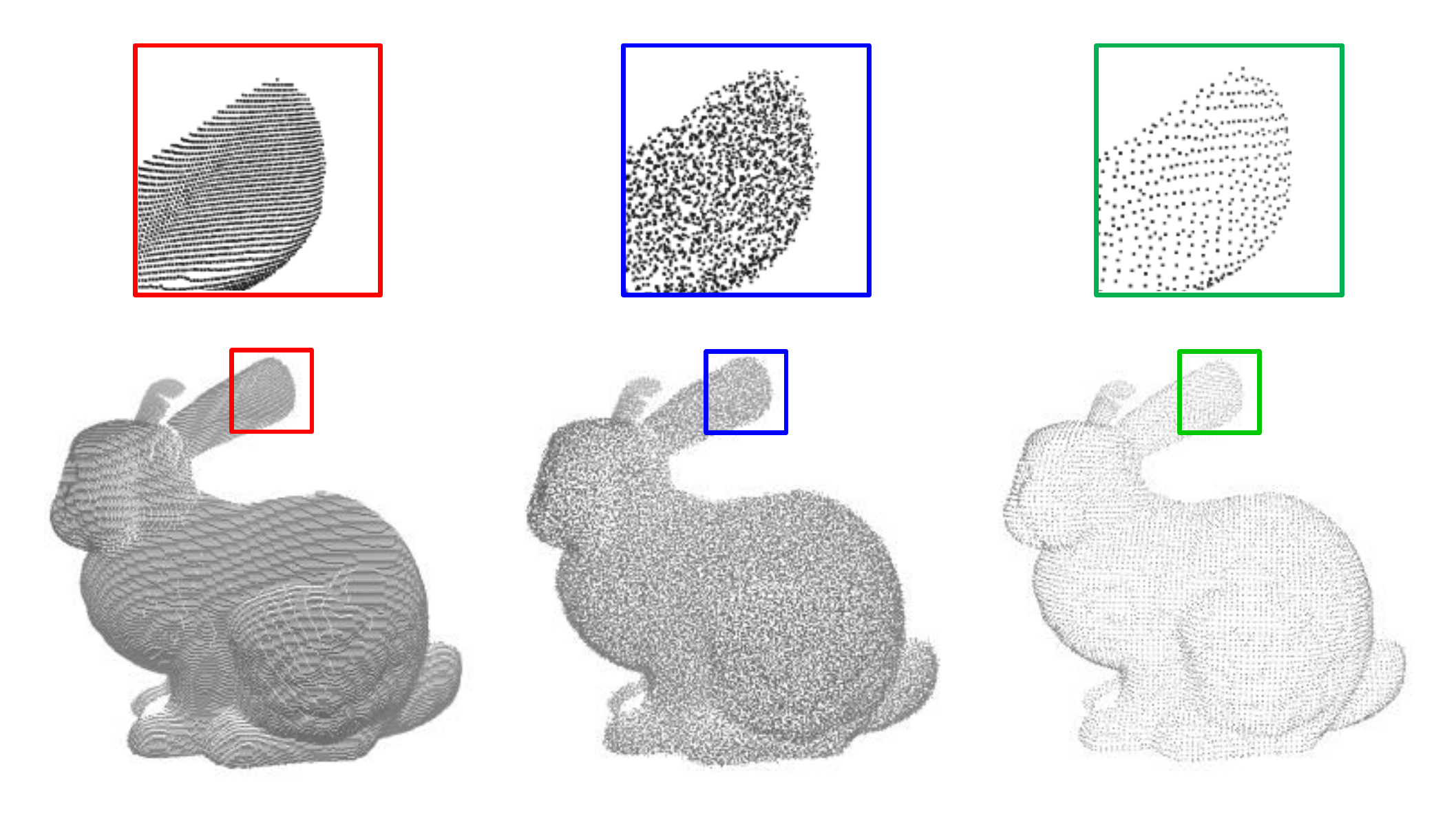}\\
	\caption{Comparison of (left) a clean rigid data and its (middle) noisy  and (right) down-sampled copies. The standard deviation of noise is 0.5 \textit{pr} and the down-sampling ratio is 0.7.}
	\label{fig:noise_density}
\end{figure}

\noindent\textbf{Noise.} Noise refer to as unwanted points near to the surface that may impair the intrinsic structure of 3D data (Fig.~\ref{fig:noise_density}). It often arises from the limitations of sensors or cluttered environments. We inject the scenes from the B3R dataset with Gaussian noise along the $x$, $y$, and $z$ axes. The standard deviation of noise increases from 0.05 \textit{pr} to 0.45 \textit{pr} with an incremental step of 0.05 \textit{pr}.
\\
\\
\noindent\textbf{Density variation.} Different from images, the change of distance from the senor to objects/scenes results in data density variation for point clouds rather than scale variations (Fig.~\ref{fig:noise_density}). This frequently happens when scanning an object from different viewpoints or monitoring a moving object. We therefore down-sample the scenes in the B3R dataset and ensure the down-sampled data undergo 90\% to 10\% data decimations with an interval of 10\%.
\begin{figure*}[t]
	\hfill
	\begin{minipage}{0.195\linewidth}
		\centering
		\subfigure[\textit{Noise}]{
			\includegraphics[width=1\linewidth]{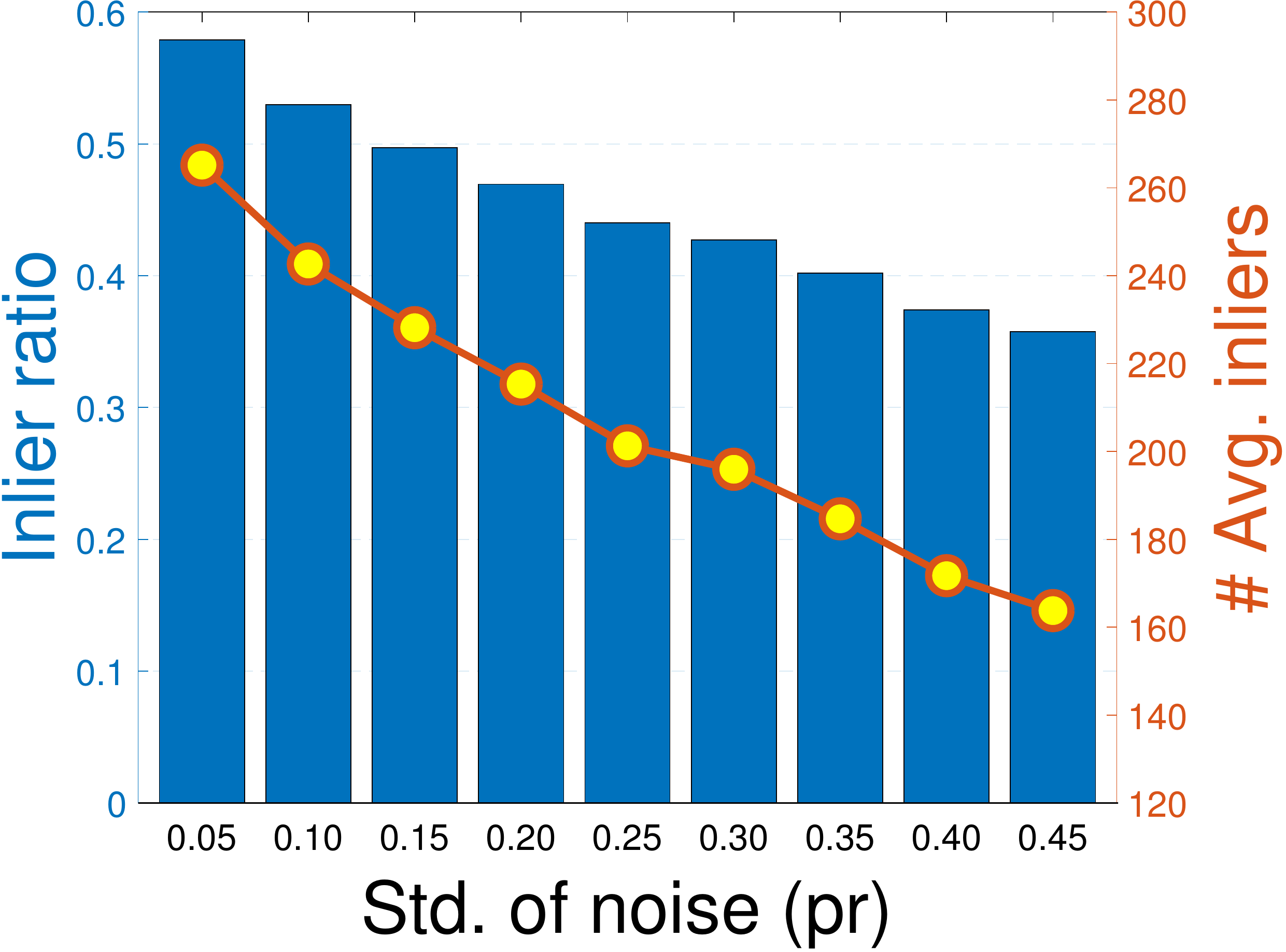}}
	\end{minipage}
	\begin{minipage}{0.195\linewidth}
		\centering
		\subfigure[\textit{Density variation}]{
			\includegraphics[width=1\linewidth]{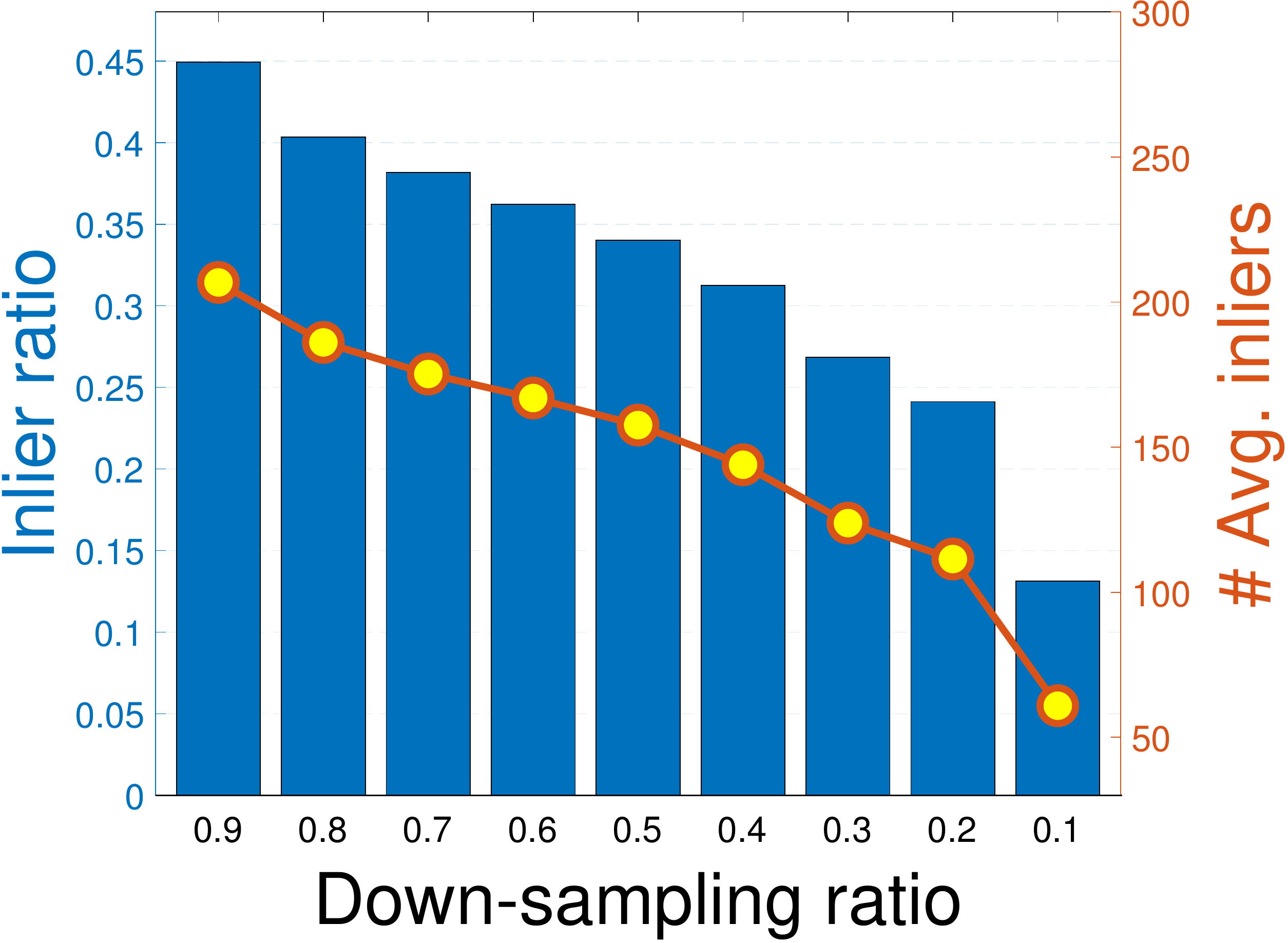}}
	\end{minipage}
	\begin{minipage}{0.195\linewidth}
		\centering
		\subfigure[\textit{Clutter}]{
			\includegraphics[width=1\linewidth]{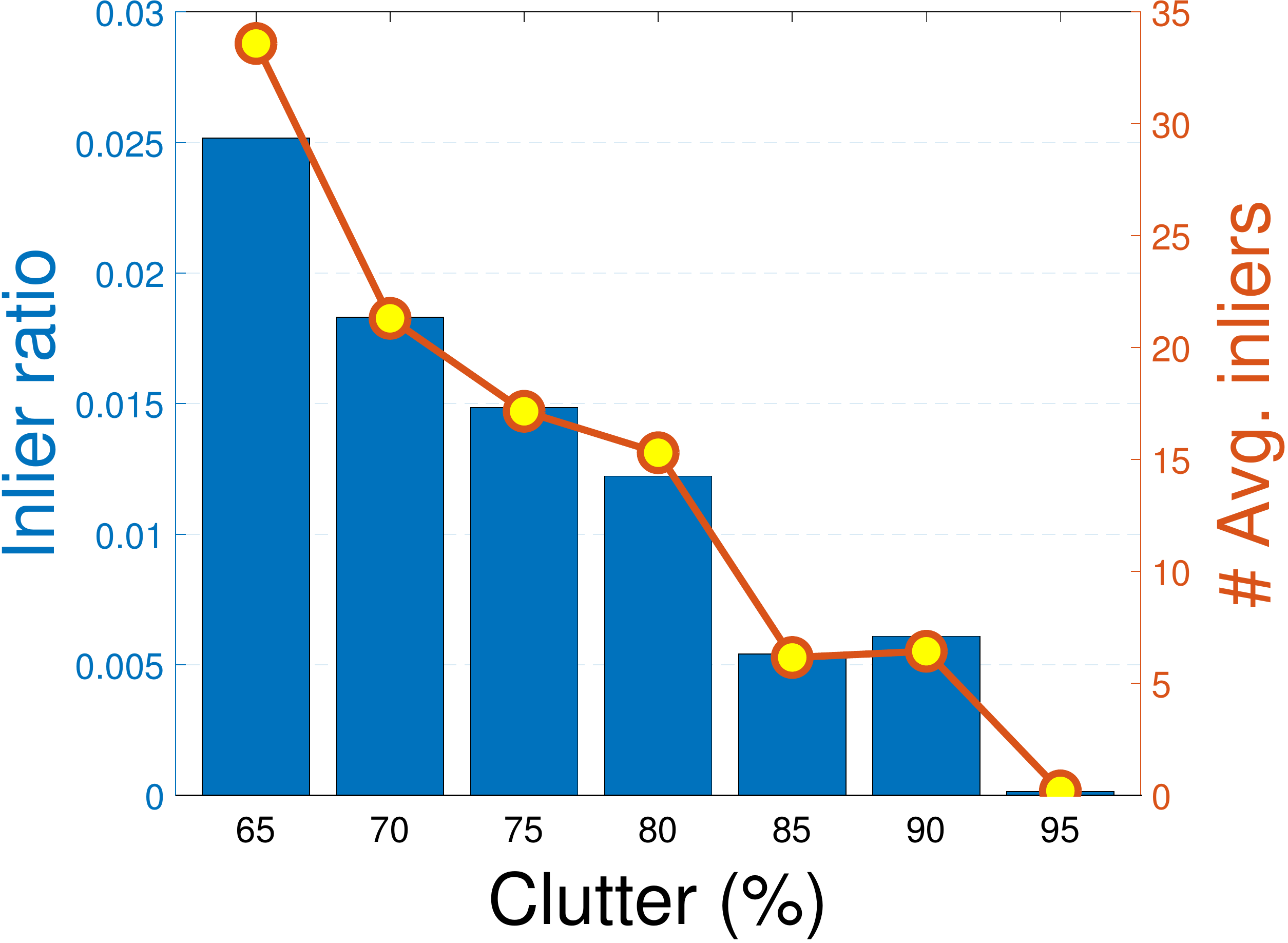}}
	\end{minipage}
	\begin{minipage}{0.195\linewidth}
		\centering
		\subfigure[\textit{Occlusion}]{
			\includegraphics[width=1\linewidth]{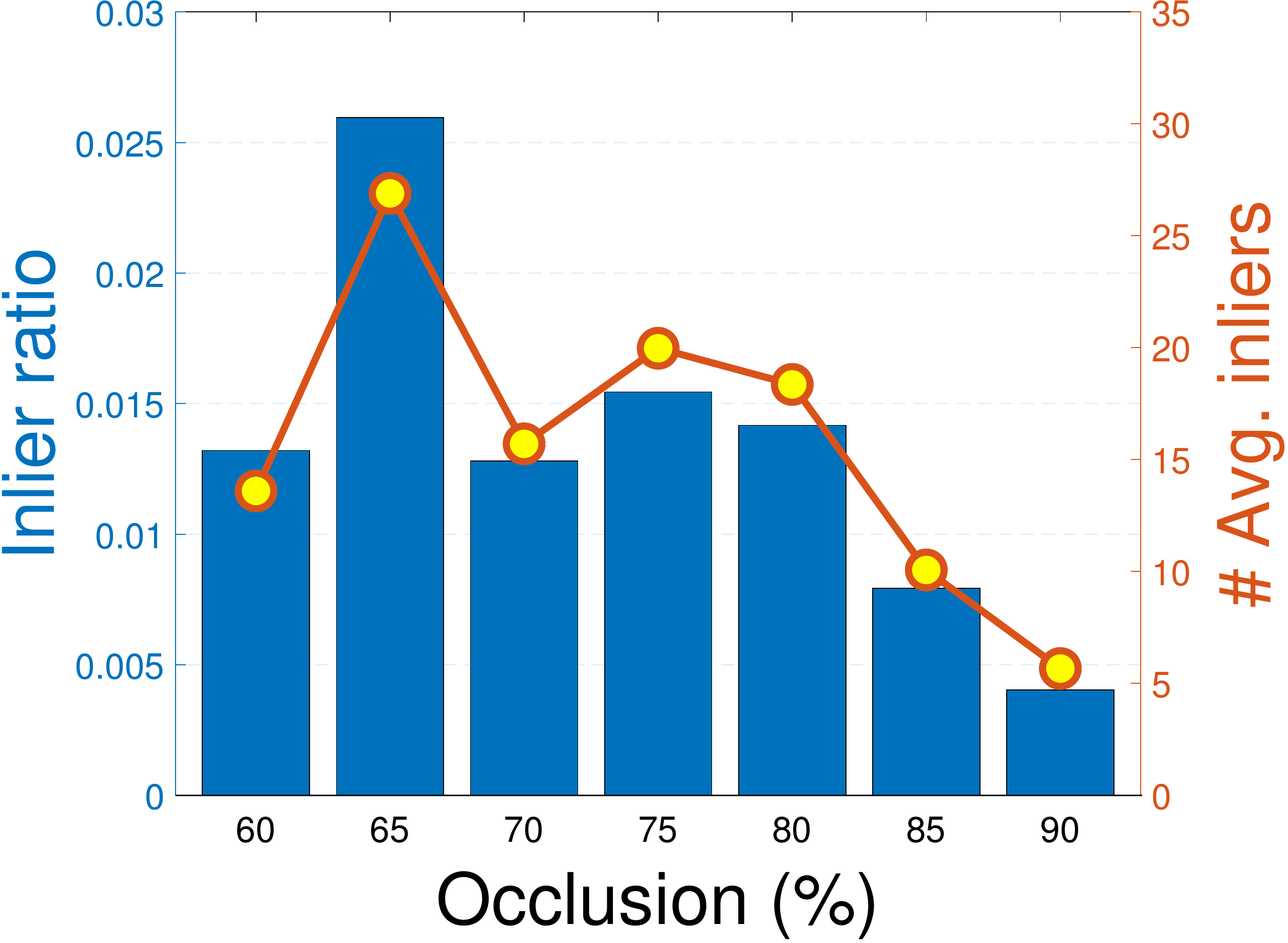}}
	\end{minipage}
	\begin{minipage}{0.195\linewidth}
		\centering
		\subfigure[\textit{Partial overlap}]{
			\includegraphics[width=1\linewidth]{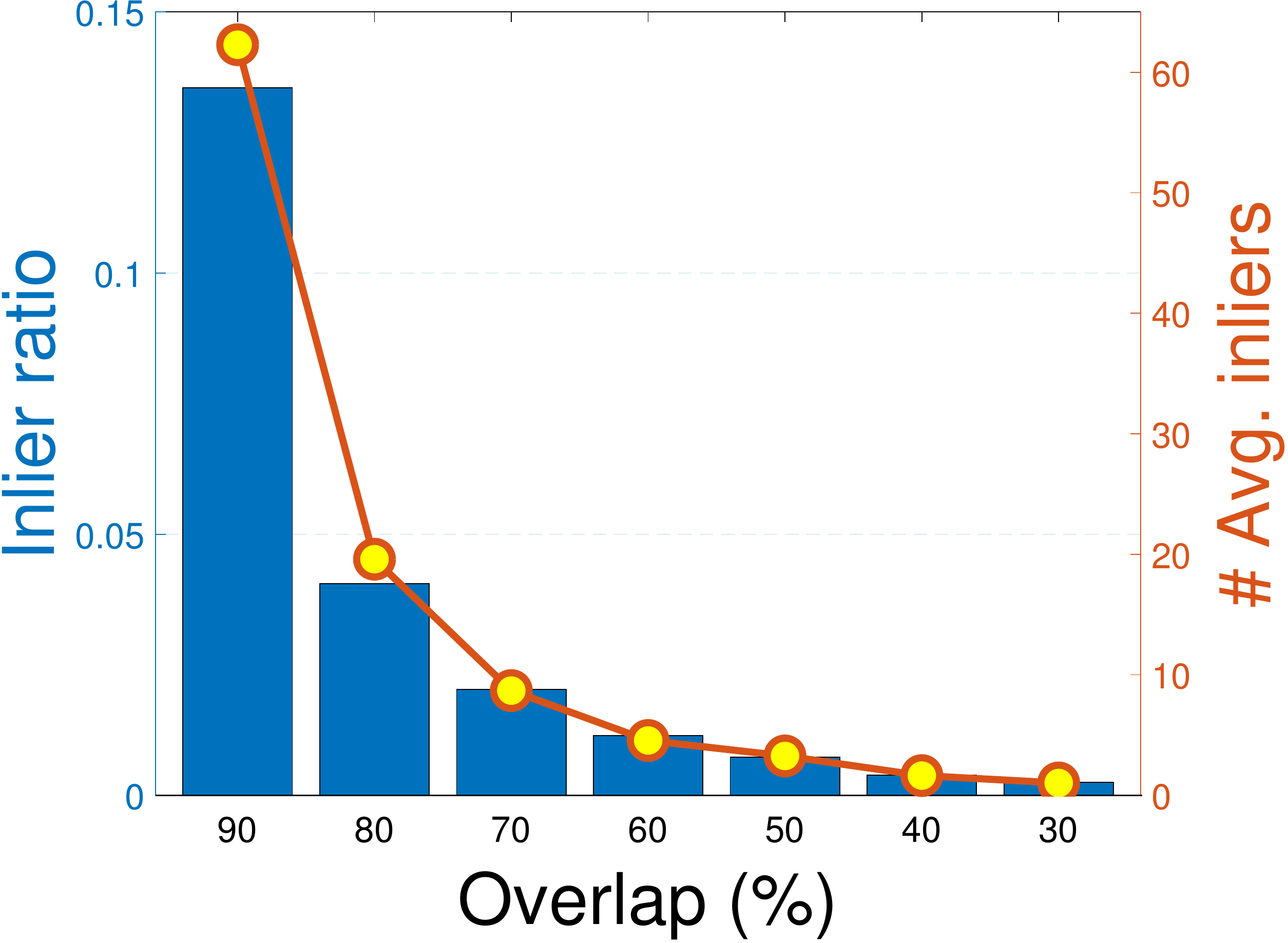}}
	\end{minipage}
	\hfill
	\begin{minipage}{0.195\linewidth}
		\centering
		\subfigure[\textit{Threshold $\epsilon$}]{
			\includegraphics[width=1\linewidth]{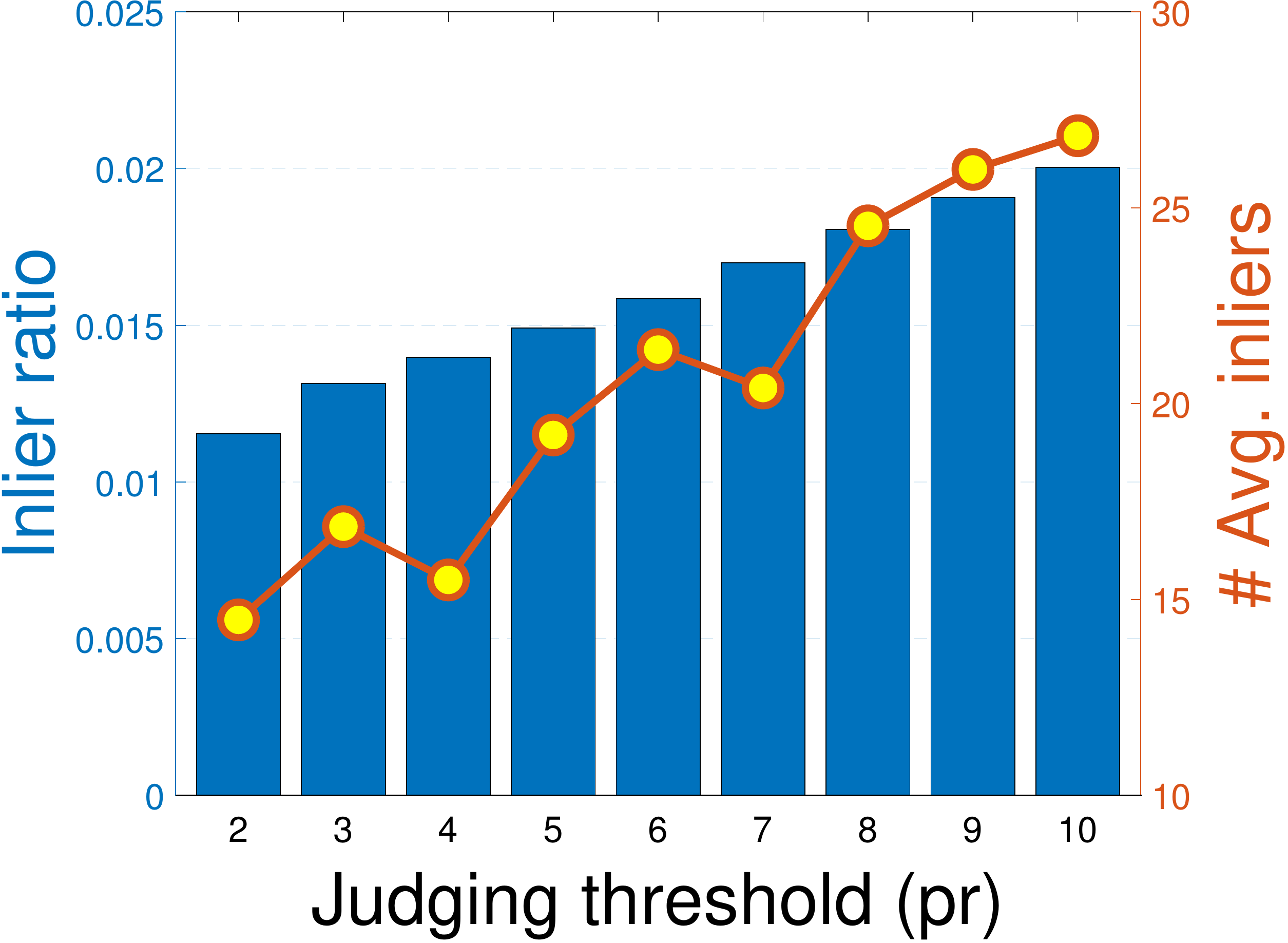}}
	\end{minipage}
	\begin{minipage}{0.195\linewidth}
		\centering
		\subfigure[\textit{Number of initial matches}]{
			\includegraphics[width=1\linewidth]{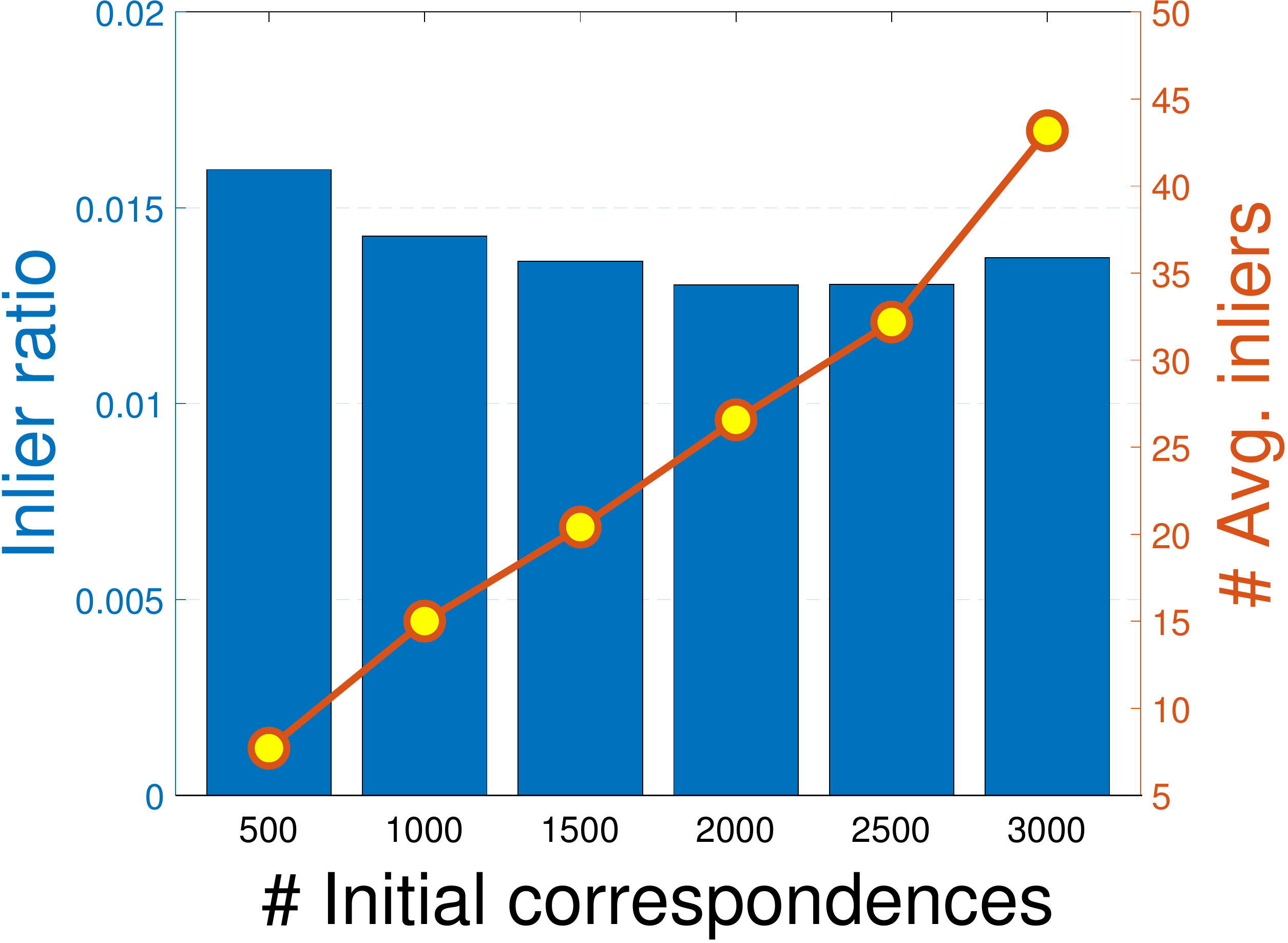}}
	\end{minipage}
	\begin{minipage}{0.195\linewidth}
		\centering
		\subfigure[\textit{Det.} \&  \textit{Desc. on B3R }]{
			\includegraphics[width=1\linewidth]{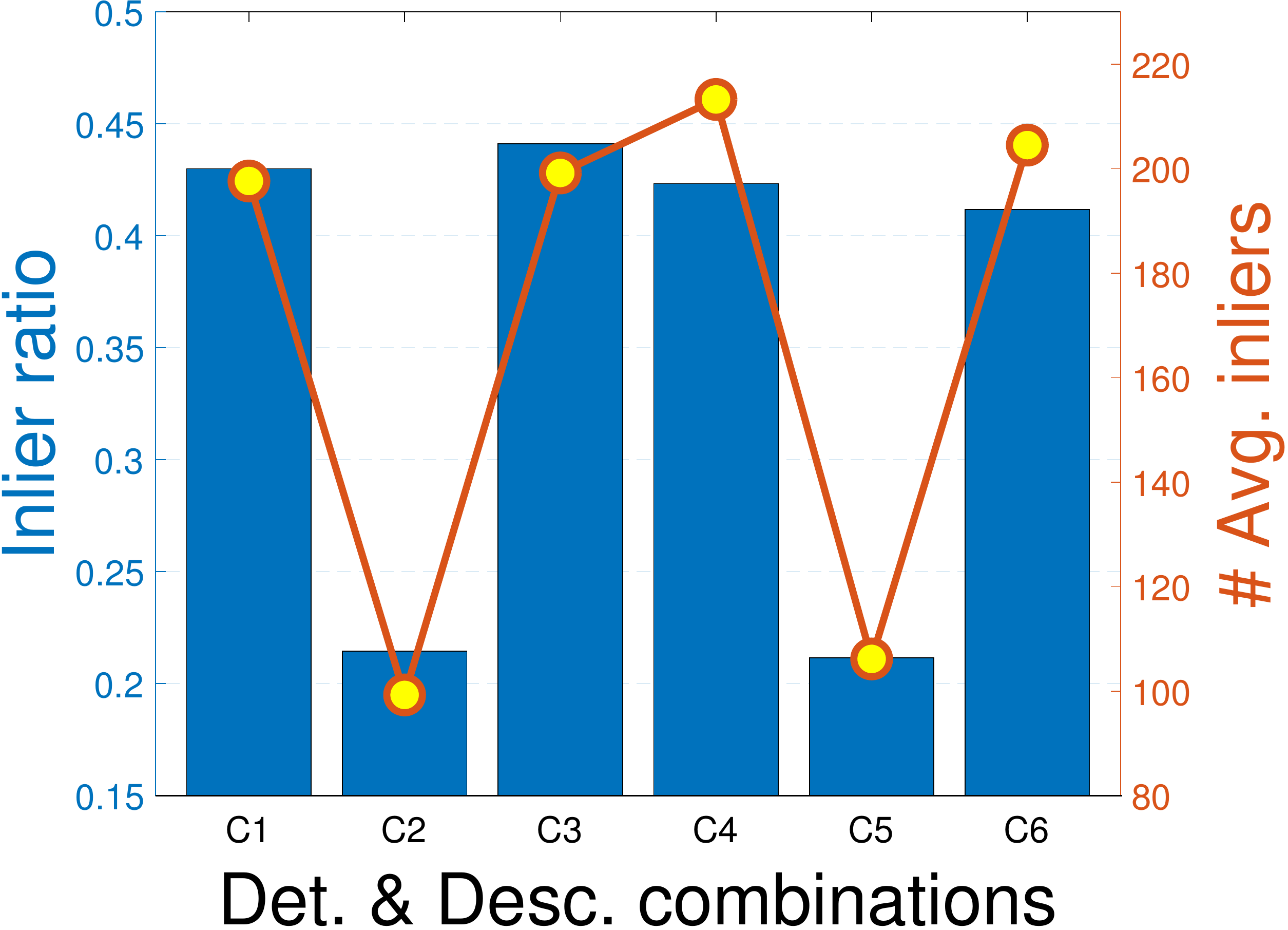}}
	\end{minipage}
	\begin{minipage}{0.195\linewidth}
		\centering
		\subfigure[\textit{Det.} \&  \textit{Desc. on U3OR }]{
			\includegraphics[width=1\linewidth]{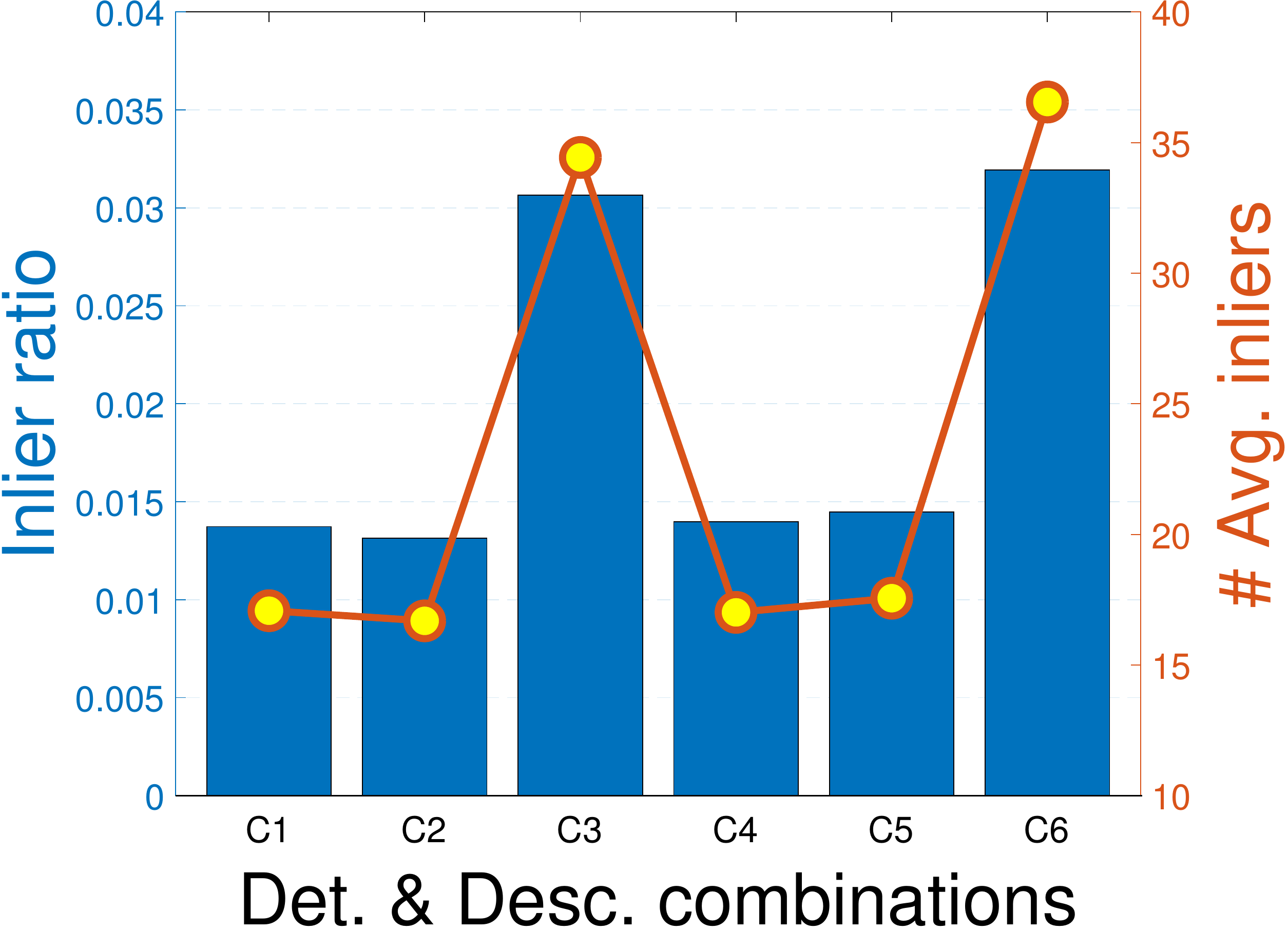}}
	\end{minipage}
	\begin{minipage}{0.195\linewidth}
		\centering
		\subfigure[\textit{Det.} \&  \textit{Desc. on U3M }]{
			\includegraphics[width=1\linewidth]{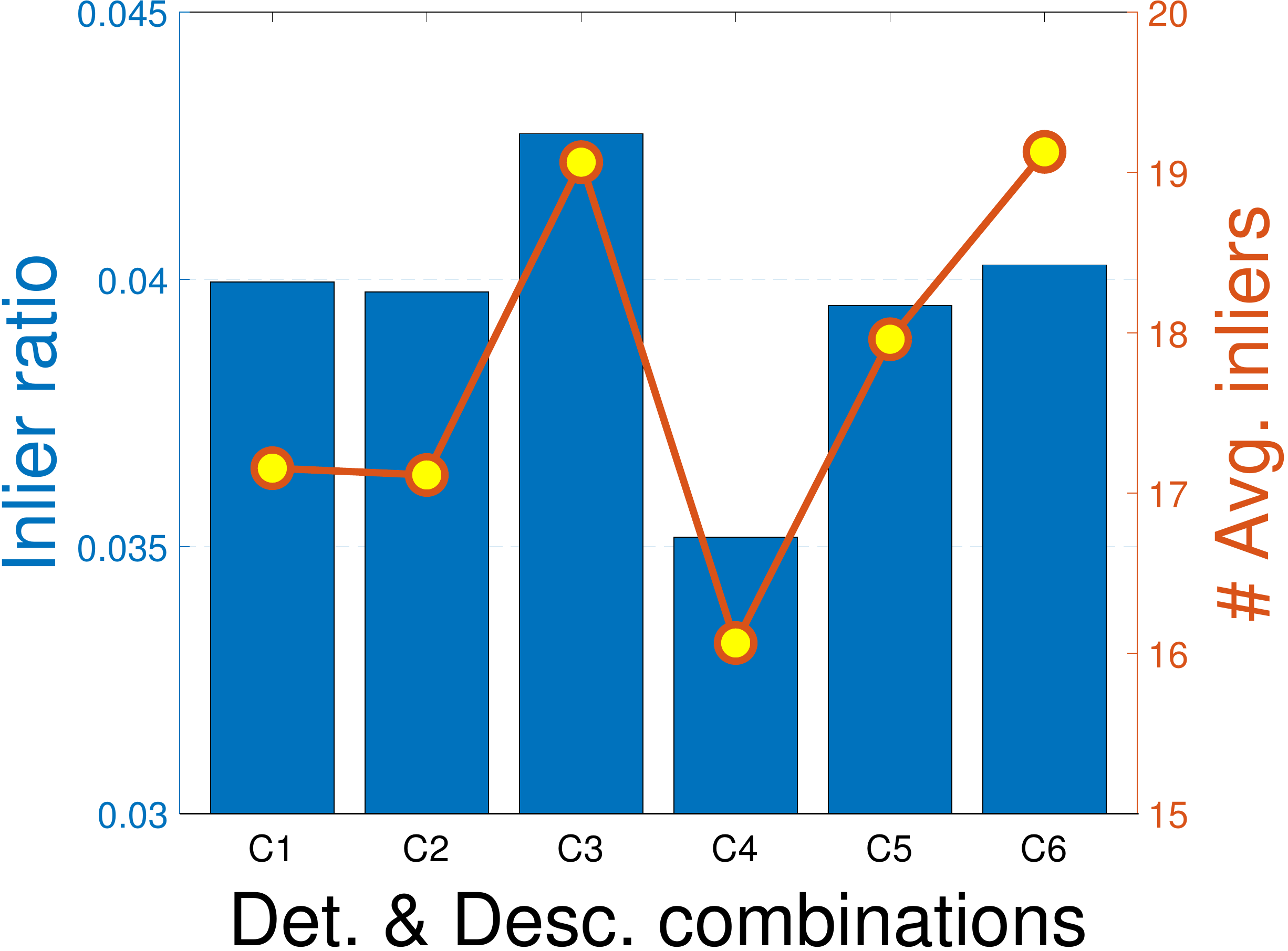}}
	\end{minipage}
	\hfill
	\caption{Information in terms of inlier ratio and the number of inliers for the initial correspondences when confronted with different challenges. Details about the generation of the initial correspondence set in each challenge case are presented in Sect.~\ref{sec:challenges}.}
	\label{fig:inlier_info}
\end{figure*}
\\
\\
\noindent\textbf{Clutter and occlusion.} For model-based 3D object recognition and pose estimation, clutter and occlusion possess great challenges for local feature matching-based methods. The U3OR dataset contains different degrees of clutter and occlusion. Specifically, clutter and occlusion are respectively defined as~\cite{mian2006three}:
\begin{equation}
\begin{aligned}
{\rm clutter}=1-\frac{{\rm model}\;{\rm surface}\;{\rm area}\;{\rm in}\;{\rm scene}}{{\rm total}\;{\rm surface}\;{\rm area}\;{\rm of}\;{\rm scene}},
\end{aligned}
\end{equation}
\begin{equation}
\begin{aligned}
{\rm occlusion}=1-\frac{{\rm model}\;{\rm surface}\;{\rm area}\;{\rm in}\;{\rm scene}}{{\rm total}\;{\rm model}\;{\rm surface}\;{\rm area}}.
\end{aligned}
\end{equation}
Accordingly, we split the U3OR dataset to 7 subsets with different degrees of clutter from 65\% to 95\% and analogously obtain another 7 subsets with different degrees of occlusion from 60\% to 90\%.
\\
\\
\noindent\textbf{Partial overlap.} Due to self-occlusion, only 2.5D data can be obtained when scanning a 3D object/scene from a particular view. The overlap ratio between two 2.5D views dominantly decides difficulty of pairwise registration~\cite{mian2006novel}. The U3M dataset has view pairs with at least 0.3 overlap. The definition of overlap is given as~\cite{mian2006novel}:
\begin{equation}
\begin{aligned}
{\rm overlap}=\frac{\#\;{\rm corr.}\;{\rm points}\;{\rm between}\;{\rm view1} \;{\rm and} \;{\rm view2}}{{\rm min}(\#\;{\rm view1}\;{\rm points},\#\;{\rm view2}\;{\rm points})}.
\end{aligned}
\end{equation} 
The U3M dataset is divided into 7 groups with different levels of overlap.
\\
\\
\noindent\textbf{Varying threshold $\epsilon$.} The threshold $\epsilon$ in Eq.~\ref{eq:judge} determines the accuracy of an inlier. As the requirement for such accuracy may vary with applications, it is necessary to evaluate the performance of a correspondence method with different values of $\epsilon$.  Specifically, we consider the whole U3OR dataset for the evaluation of this term. Note that we set $\epsilon$ to 5 \textit{pr} by default.
\\
\\
\noindent\textbf{Varying numbers of initial matches.} Different numbers of initial matches are desired respecting different tasks, e.g., \textit{dense} matching for shape morphing~\cite{alexa2002recent} and \textit{sparse} matching for crude scan alignment. The number of initial matches is controlled by the number of detected keypoints on both data. In traditional 3D keypoint detectors, the sparsity of detected keypoints is usually determined by a parameter called non-maximum-suppression (NMS) radius~\cite{tombari2013performance}. We vary the NMS radius of the employed keypoint detector to generate initial matches with different sizes.
\\
\\
\noindent\textbf{Different detector-descriptor combinations.} The spatial distribution and quality of initial correspondences can be also affected by the chosen keypoint detector and descriptor. In our evaluation, we consider two popular 3D keypoint detectors, i.e., Harris 3D (H3D)~\cite{sipiran2011harris} and intrinsic shape signatures (ISS)~\cite{zhong2009intrinsic}, and three representative local geometric descriptors, i.e., SHOT~\cite{tombari2010unique}, local feature statistics histograms (LFSH)~\cite{yang2016fast}, and rotational contour signatures (RCS)~\cite{yang2017RCS_jrnl}. All possible combinations of these detectors and descriptors are considered. In the following, we use ``C1$\sim$C6'' to represent the combinations of H3D+SHOT, H3D+LFSH, H3D+RCS, ISS+SHOT, ISS+LFSH, and ISS+RCS, respectively. This term is tested on all experimental datasets.  When testing the robustness to other nuisances, we employ H3D+SHOT to generate raw matches by default.

\begin{table}[t]\small
	\renewcommand{\arraystretch}{1}
	\caption{Parameters used through the evaluation.}
	\label{tab:para}
	\centering
	\begin{tabular}{@{\extracolsep{\fill}}lll}
		\hline
		SS  & $t_{ss}$ & Adaptive~\cite{otsu1975threshold} \\
		\hline
		NNSR~\cite{lowe2004distinctive} & $t_{nnsr}$ & 0.8 \\
		\hline
		RANSAC~\cite{fischler1981random} & $N_{ransac}$ & 10000 \\
		& $d_{ransac}$ & 5 \textit{pr}\\
		\hline
		ST~\cite{leordeanu2005spectral} & $t_{st}$ & 0.6\\
		\hline
		GC~\cite{chen20073d} & $t_{gc}$ & 3 \textit{pr}\\ 
		\hline
		3DHV~\cite{tombari2010object} & - &- \\
		\hline
		GTM~\cite{rodola2013scale} & $N_{gtm}$&100\\
		& $t_{gtm}$&Adaptive~\cite{otsu1975threshold}\\
		\hline
		SI~\cite{buch2014search} & $\kappa$ & 250 \\
		& $\varsigma$ & 0.9 \\
		& $\delta$ & 5 \textit{pr}\\
		\hline
		CV~\cite{yang2019ranking}&$k$&200\\
		&$t_{cv}$&Adaptive~\cite{otsu1975threshold} \\
		\hline
	\end{tabular} 
\end{table}
In order to examine the impact of above challenges on the resulted correspondences and  improve the interpretability for the experimental results as will be shown in Sect.~\ref{sec:result}, we present the results respecting the  number of inliers and inler ratio for each challenge case in Fig.~\ref{fig:inlier_info}. As expected, the rich variety of application scenarios and perturbations produces a number of correspondence sets with different qualities. For instance, correspondences computed on the B3R dataset with 0.05 \textit{pr} Gaussian noise have more than 50\% inliers while these computed on the U3OR dataset generally have less than 3\% inliers.

\subsection{Implementation Details}
The input for the tested methods in this paper, i.e., the initial correspondence set $\cal C$, is generated by matching the $L_2$ distance of keypoint descriptors extracted from the source and target shapes using $k$d-tree. The support radius of employed local geometric descriptor is set to 15 \textit{pr}~\cite{guo2013rotational}. 

As for the parameter settings of evaluated approaches, we report them in Table~\ref{tab:para}. For SS, GTM, and CV, we make the thresholds used for splitting correspondence set adaptive as suggested by~\cite{buch2014search}. For ST and GC, their thresholds are determined via tuning experiments. For RANSAC, 10000 loops are assigned to strike a balance between accuracy and efficiency. The remaining parameters are kept consistent to the settings in their original papers.

All considered methods are implemented in the point cloud library (PCL)~\cite{rusu20113d} with a 3.4GHz processor and 24GB RAM.

\section{Results}\label{sec:result}
Based on the experimental setup described in the previous section, this section presents corresponding outcomes with necessary explanations and discussions.
\subsection{Correspondence Grouping Performance}
\begin{figure*}[htbp]
	\begin{minipage}{0.194\linewidth}
		\centering
		\subfigure[\textit{Noise}]{
			\includegraphics[width=1\linewidth]{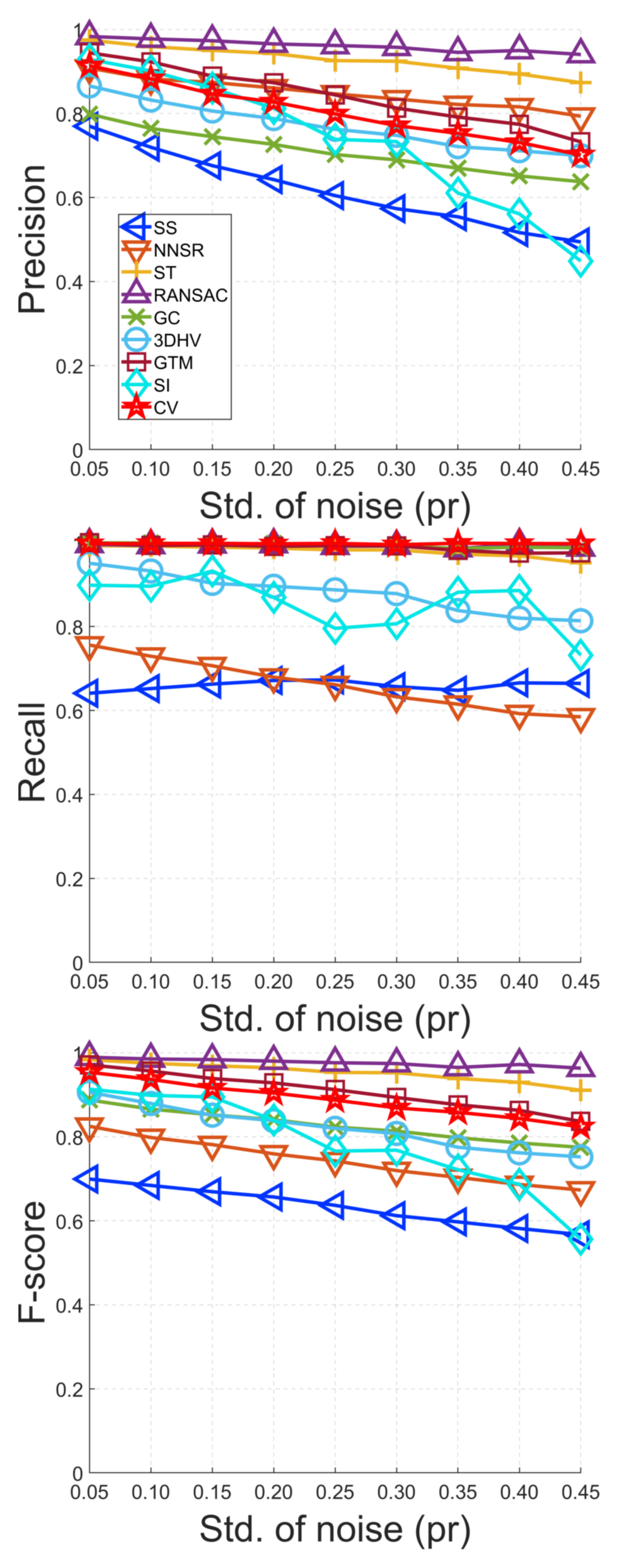}}
	\end{minipage}
	\begin{minipage}{0.194\linewidth}
		\centering
		\subfigure[\textit{Density variation}]{
			\includegraphics[width=1\linewidth]{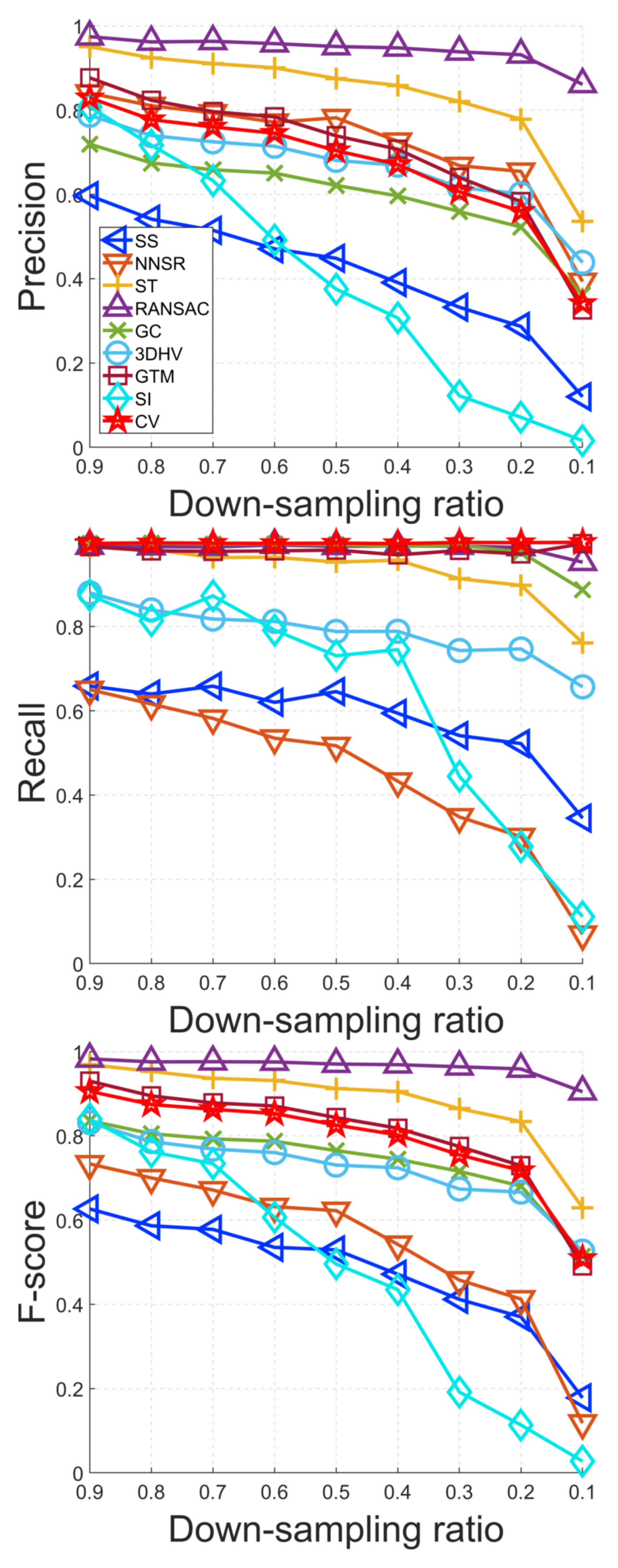}}
	\end{minipage}
	\begin{minipage}{0.194\linewidth}
		\centering
		\subfigure[\textit{Clutter}]{
			\includegraphics[width=1\linewidth]{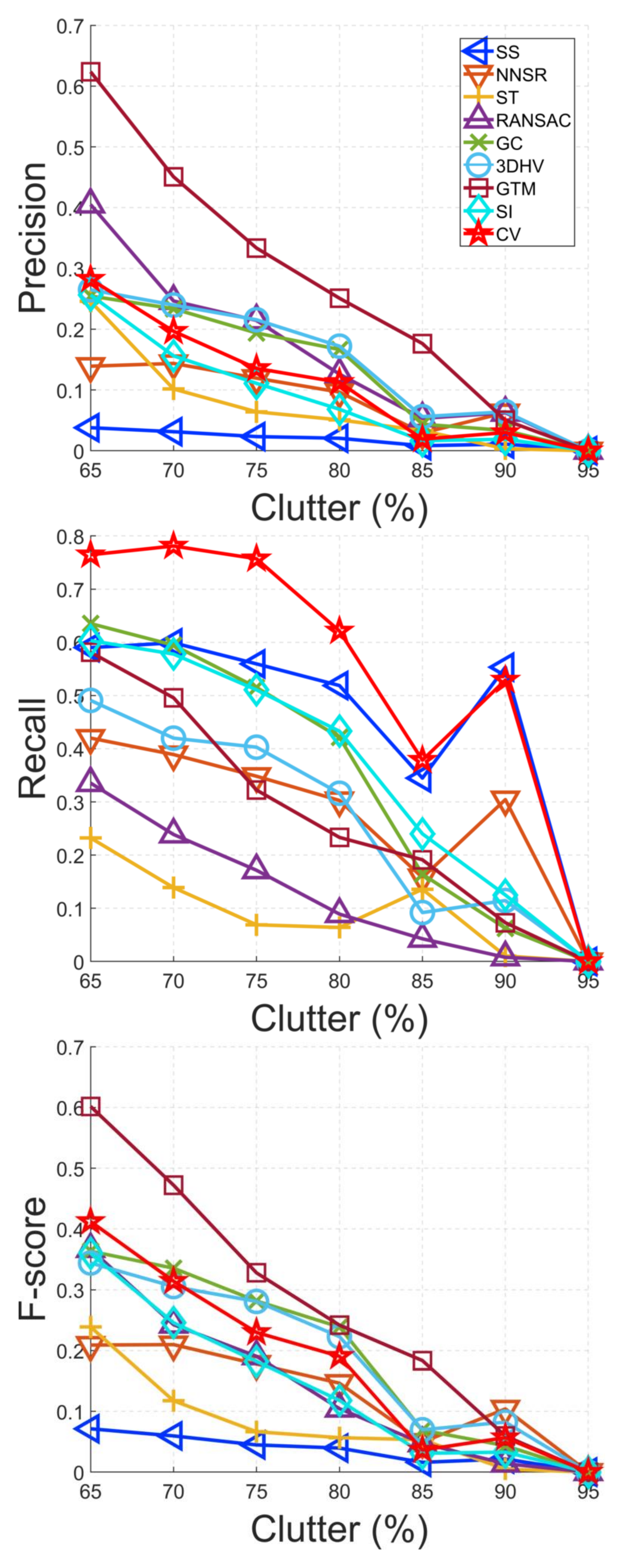}}
	\end{minipage}
	\begin{minipage}{0.194\linewidth}
		\centering
		\subfigure[\textit{Occlusion}]{
			\includegraphics[width=1\linewidth]{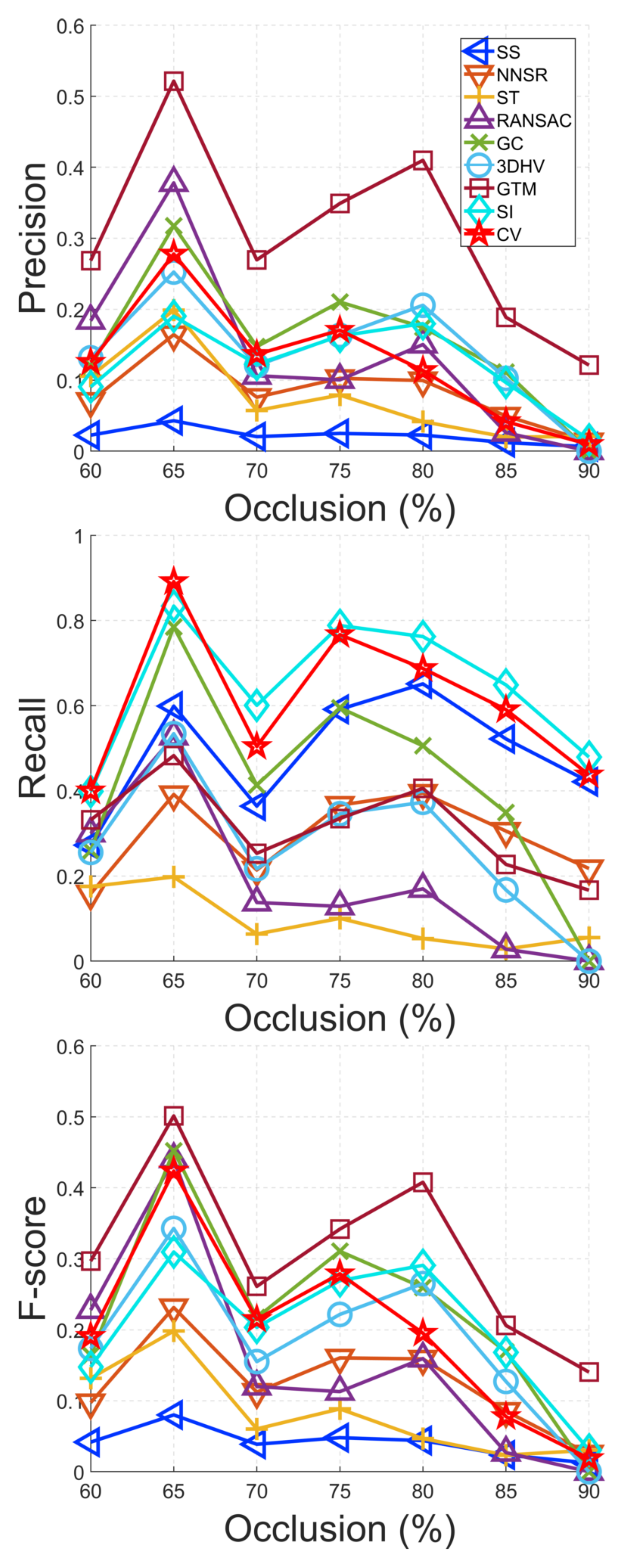}}
	\end{minipage}
	\begin{minipage}{0.194\linewidth}
		\centering
		\subfigure[\textit{Partial overlap}]{
			\includegraphics[width=1\linewidth]{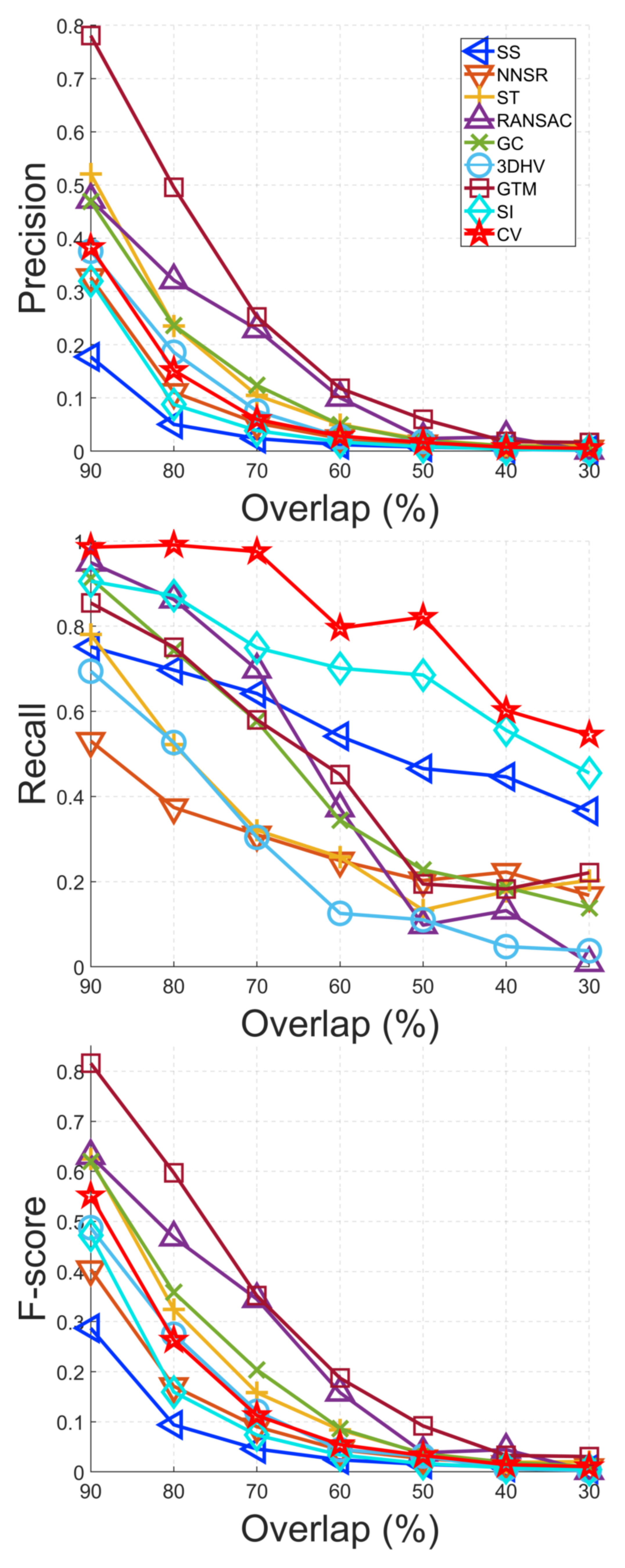}}
	\end{minipage}
	\begin{minipage}{0.194\linewidth}
		\centering
		\subfigure[\textit{Threshold $\epsilon$}]{
			\includegraphics[width=1\linewidth]{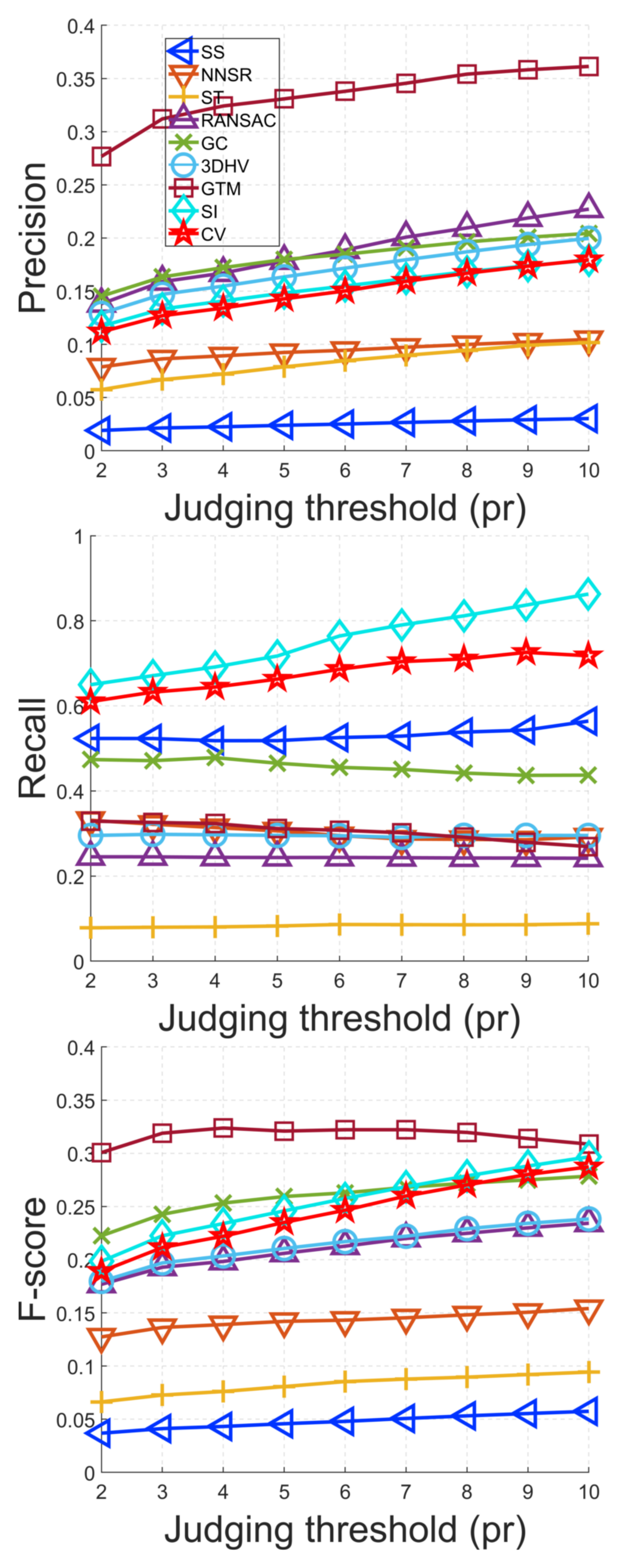}}
	\end{minipage}
	\begin{minipage}{0.194\linewidth}
		\centering
		\subfigure[\textit{Number of initial matches}]{
			\includegraphics[width=1\linewidth]{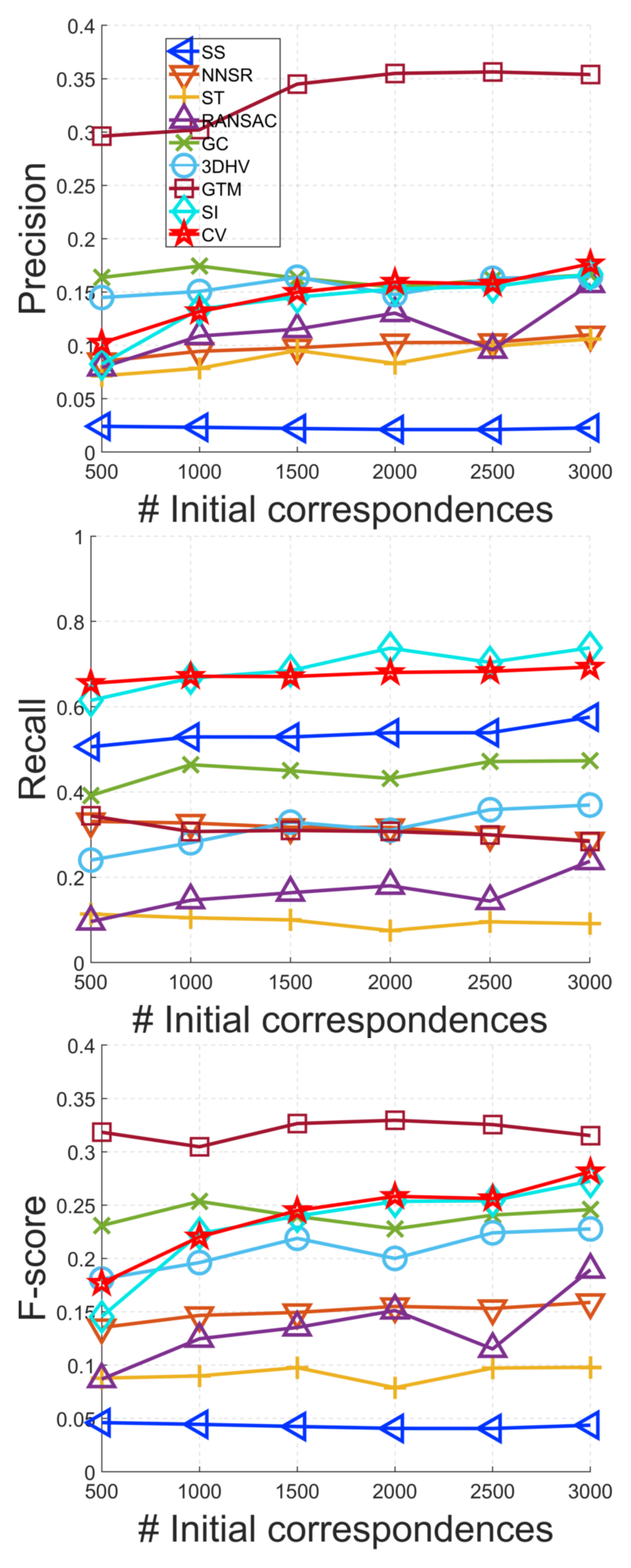}}
	\end{minipage}
	\begin{minipage}{0.194\linewidth}
		\centering
		\subfigure[\textit{Det.} \&  \textit{Desc. on B3R }]{
			\includegraphics[width=1\linewidth]{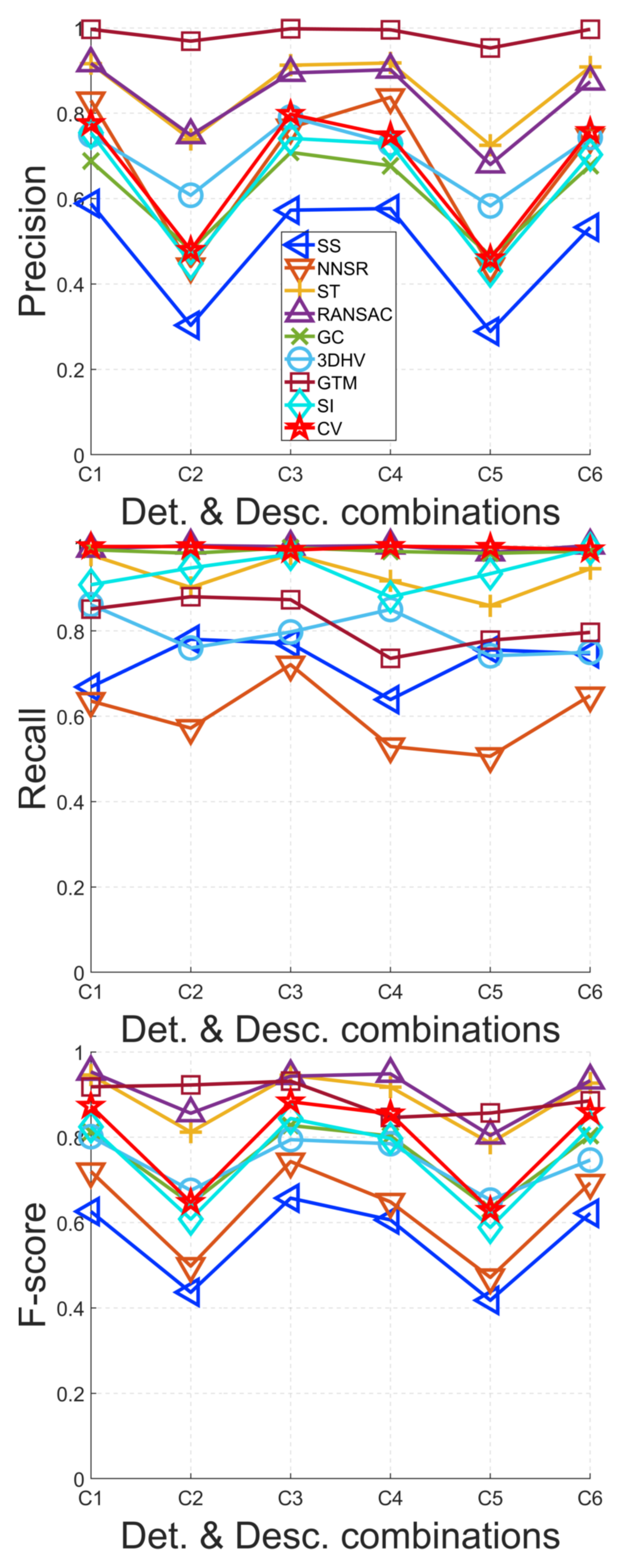}}
	\end{minipage}
	\begin{minipage}{0.194\linewidth}
		\centering
		\subfigure[\textit{Det.} \&  \textit{Desc. on U3OR }]{
			\includegraphics[width=1\linewidth]{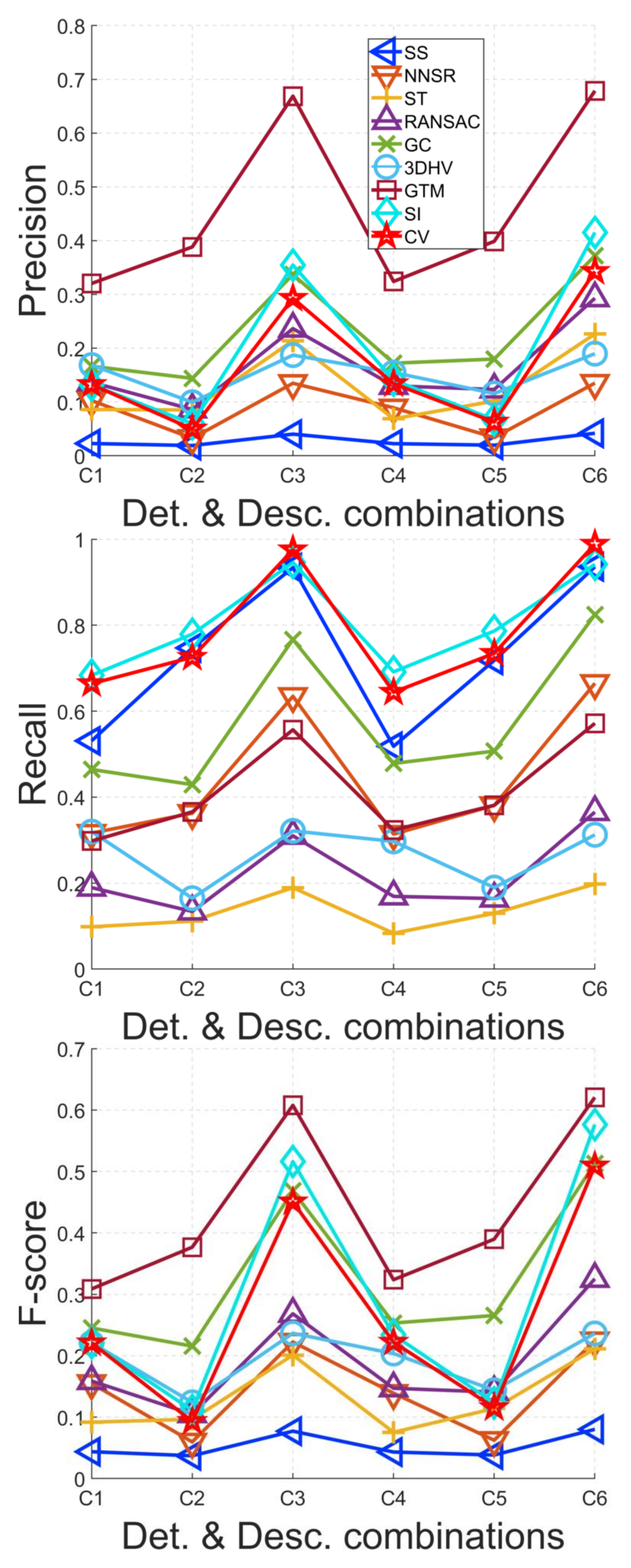}}
	\end{minipage}
	\begin{minipage}{0.194\linewidth}
		\centering
		\subfigure[\textit{Det.} \&  \textit{Desc. on U3M }]{
			\includegraphics[width=1\linewidth]{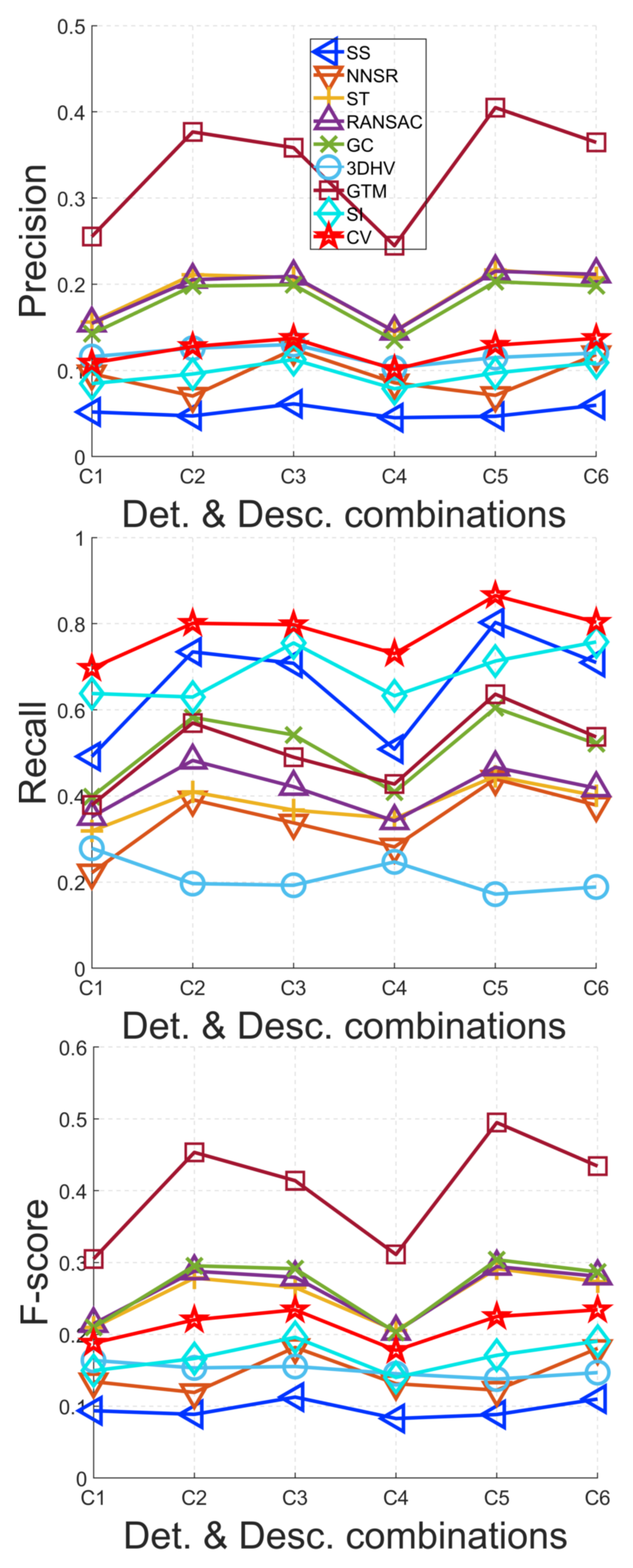}}
	\end{minipage}
	\caption{Precision, recall, and F-score performance of selected methods in Sect.~\ref{sec:eval_mtd} on experimental datasets with different challenges  (Sect.~\ref{sec:challenges}).}
	\label{fig:PRF}
\end{figure*}
The results of selected methods in Sect.~\ref{sec:eval_mtd} in terms of precision, recall, and F-score are aggregatedly presented in Fig.~\ref{fig:PRF}. Generally, the performance of each method changes frequently when faced with different nuisances. Specifically, we could make the following observations.
\begin{figure*}[htbp]
	\begin{minipage}{0.47\linewidth}
		\raggedleft
		\subfigure[\textit{Noise}]{
			\includegraphics[width=0.85\linewidth]{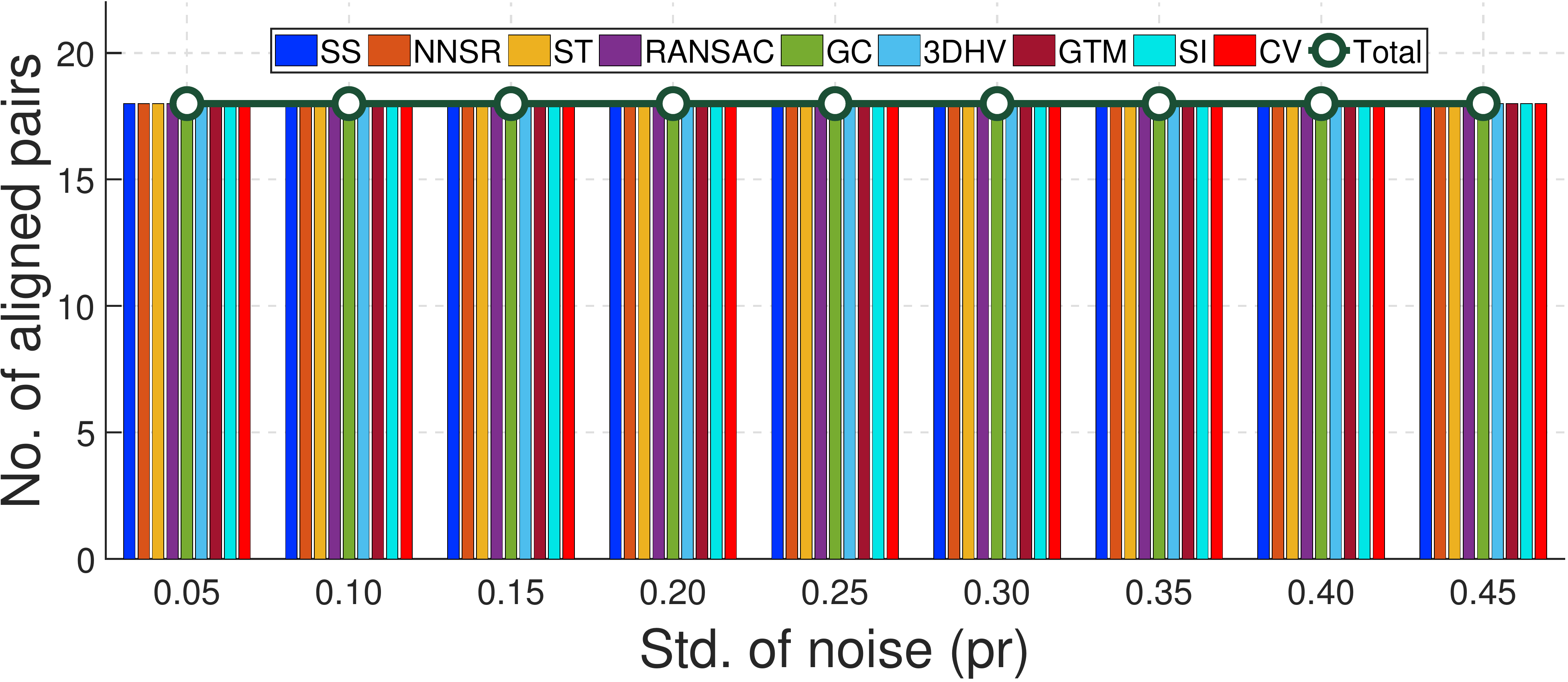}}
	\end{minipage}
	\begin{minipage}{0.47\linewidth}
		\raggedright
		\subfigure[\textit{Density variation}]{
			\includegraphics[width=0.85\linewidth]{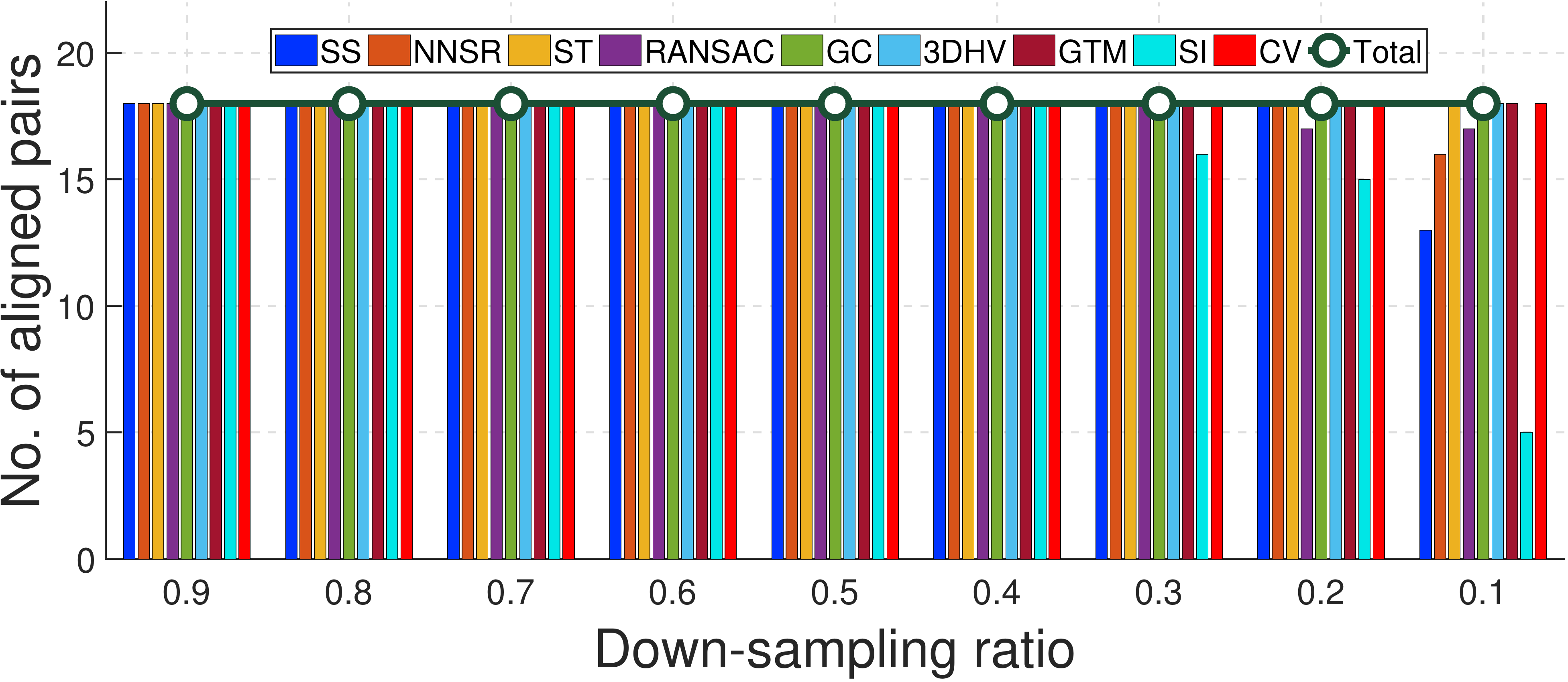}}
	\end{minipage}
	\begin{minipage}{0.47\linewidth}
		\raggedleft
		\subfigure[\textit{Clutter}]{
			\includegraphics[width=0.85\linewidth]{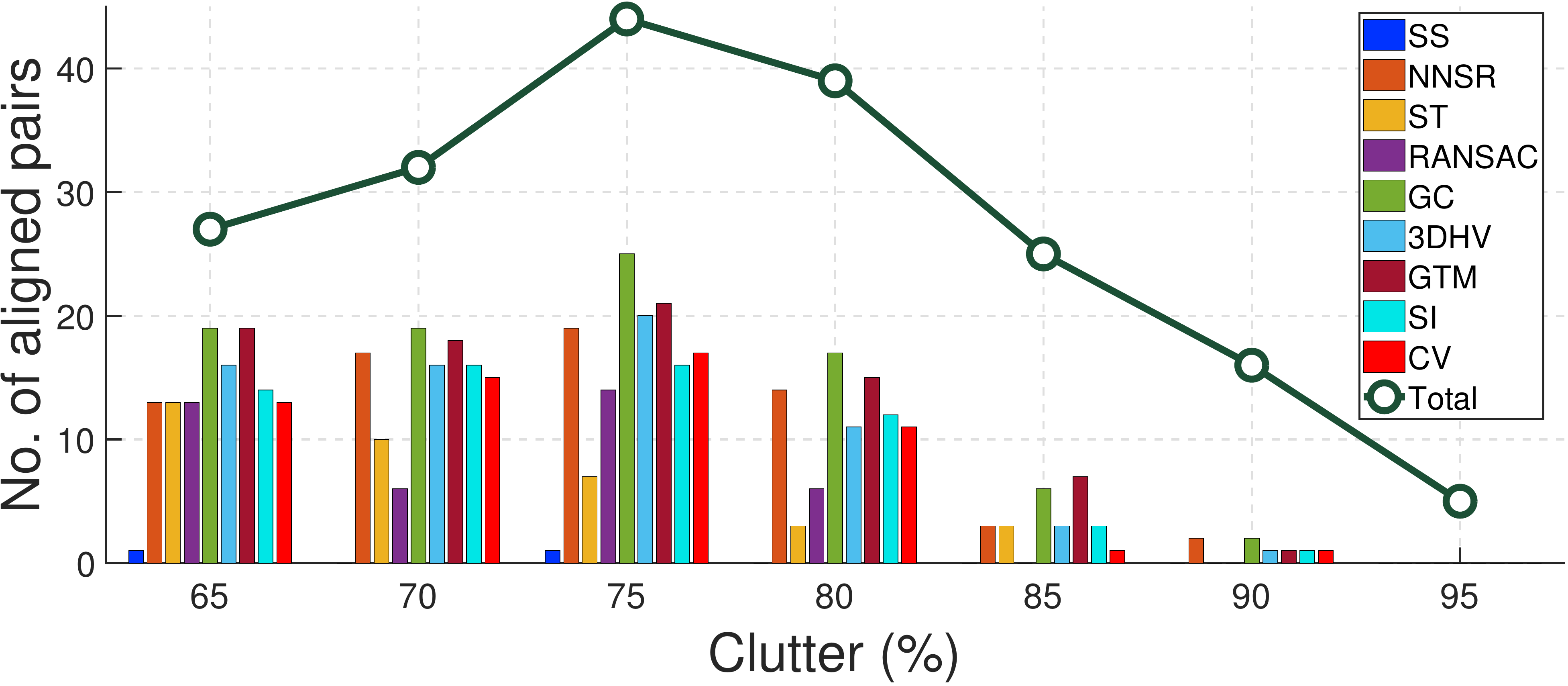}}
	\end{minipage}
	\begin{minipage}{0.47\linewidth}
		\raggedright
		\subfigure[\textit{Occlusion}]{
			\includegraphics[width=0.85\linewidth]{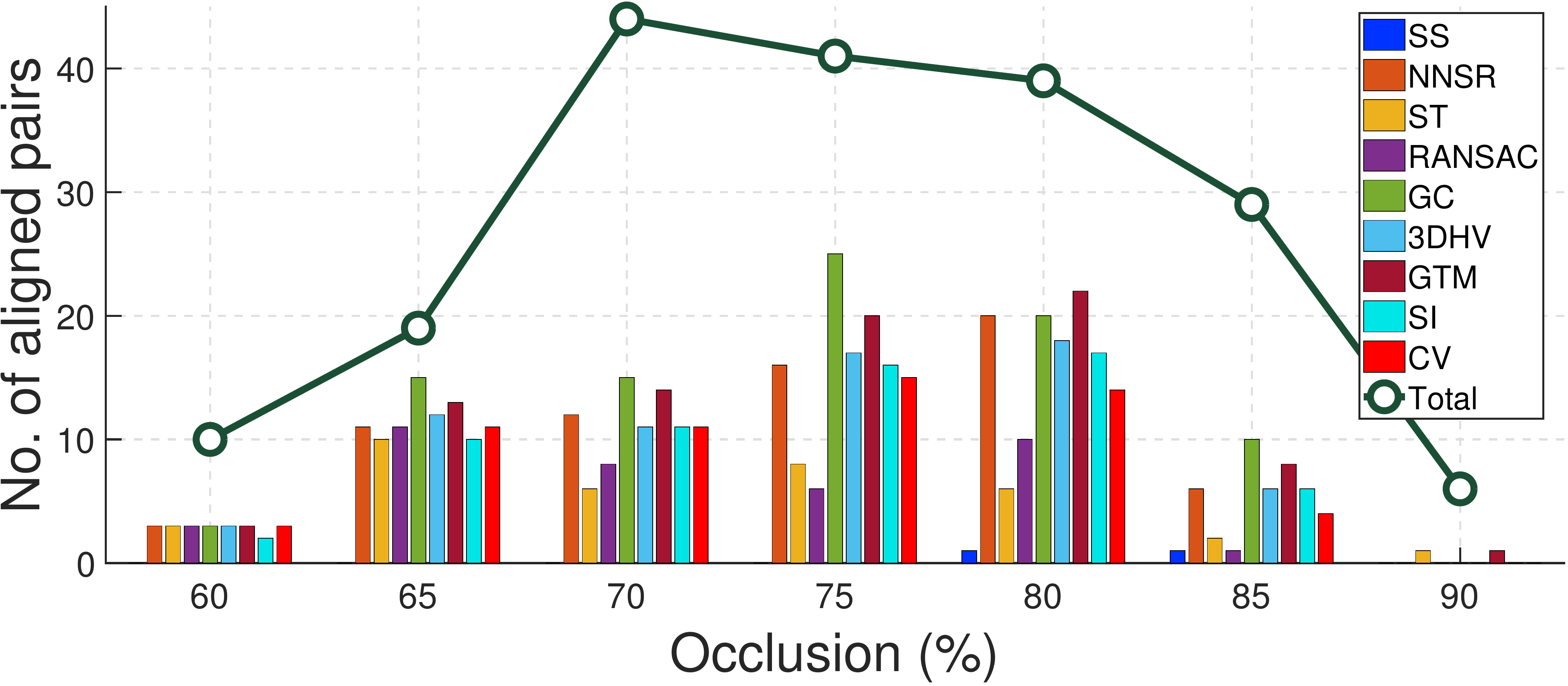}}
	\end{minipage}
	\begin{minipage}{0.47\linewidth}
		\raggedleft
		\subfigure[\textit{Partial overlap}]{
			\includegraphics[width=0.85\linewidth]{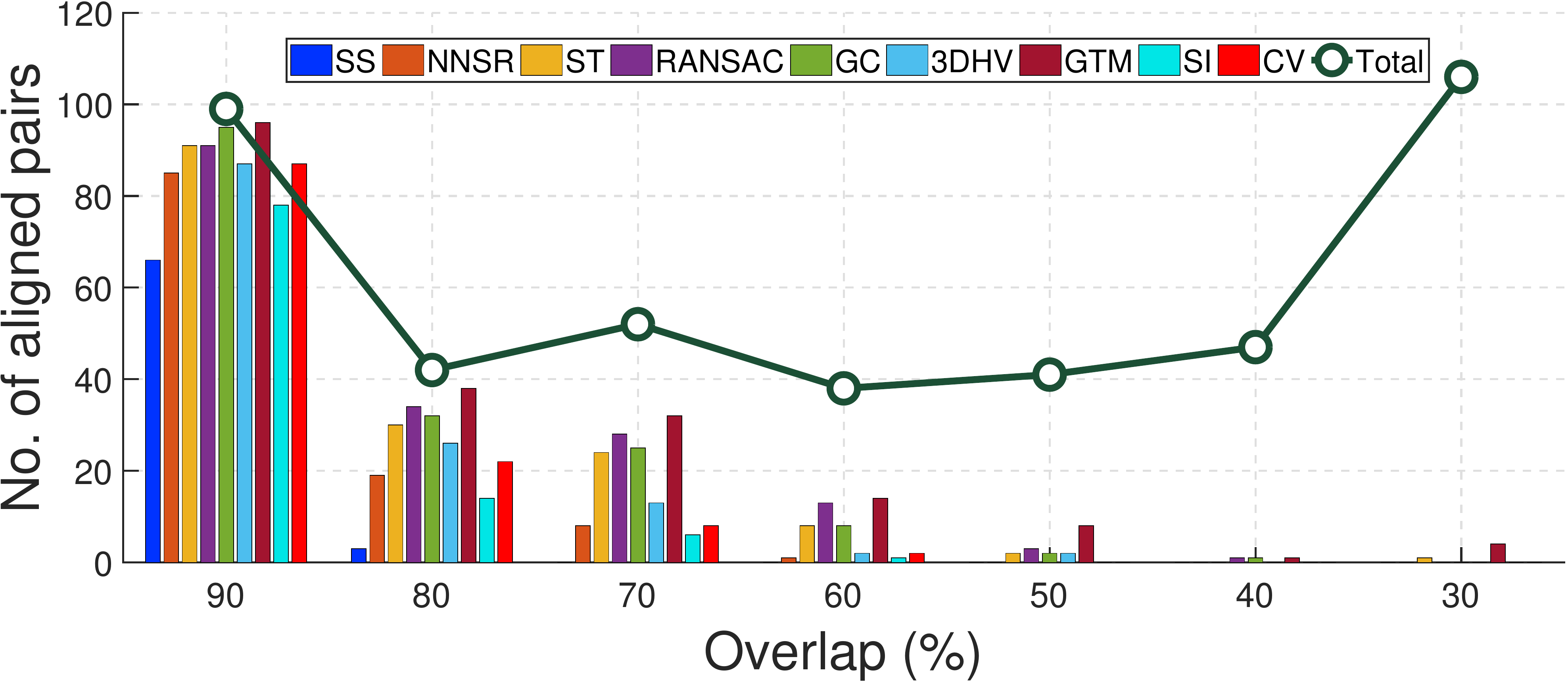}}
	\end{minipage}
	\begin{minipage}{0.47\linewidth}
		\raggedright
		\subfigure[\textit{Threshold $\epsilon$}]{
			\includegraphics[width=0.85\linewidth]{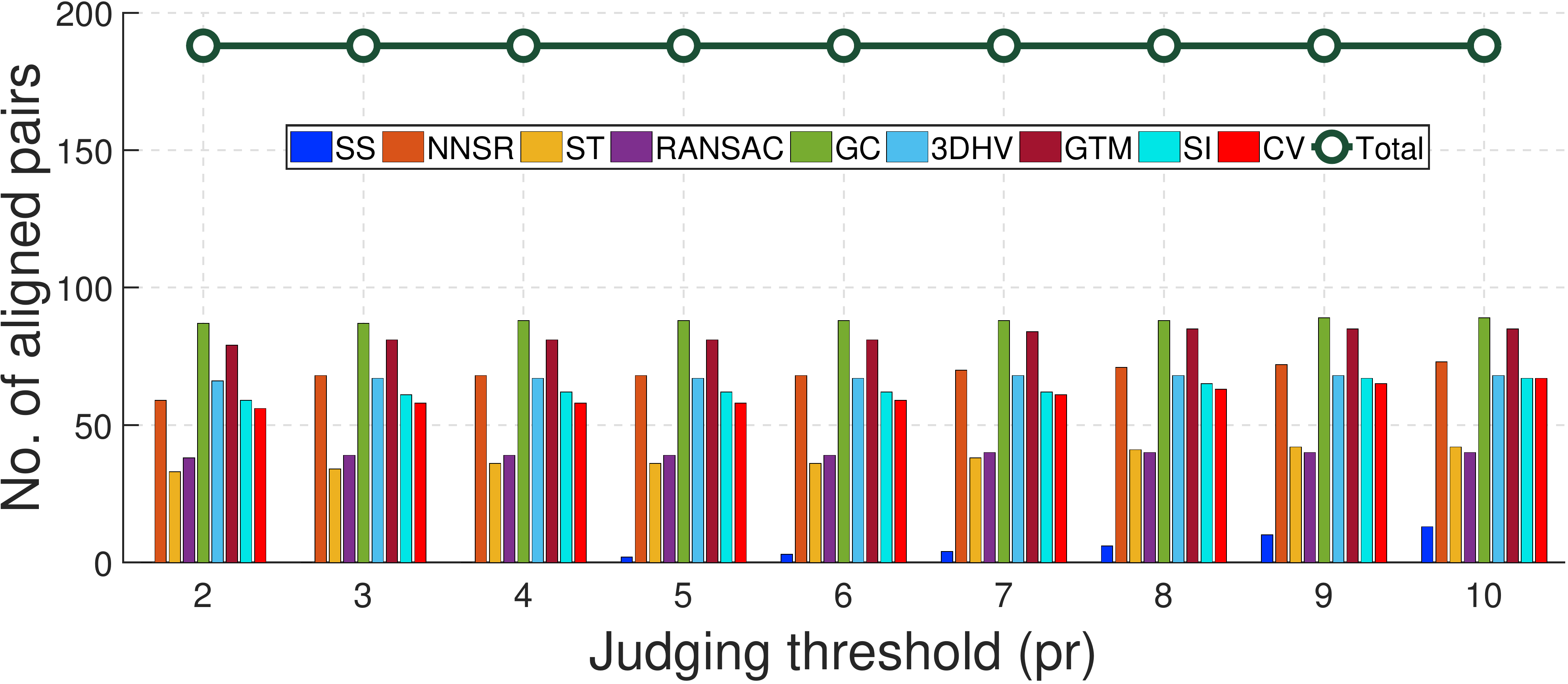}}
	\end{minipage}
	\begin{minipage}{0.47\linewidth}
		\raggedleft
		\subfigure[\textit{Number of initial matches}]{
			\includegraphics[width=0.85\linewidth]{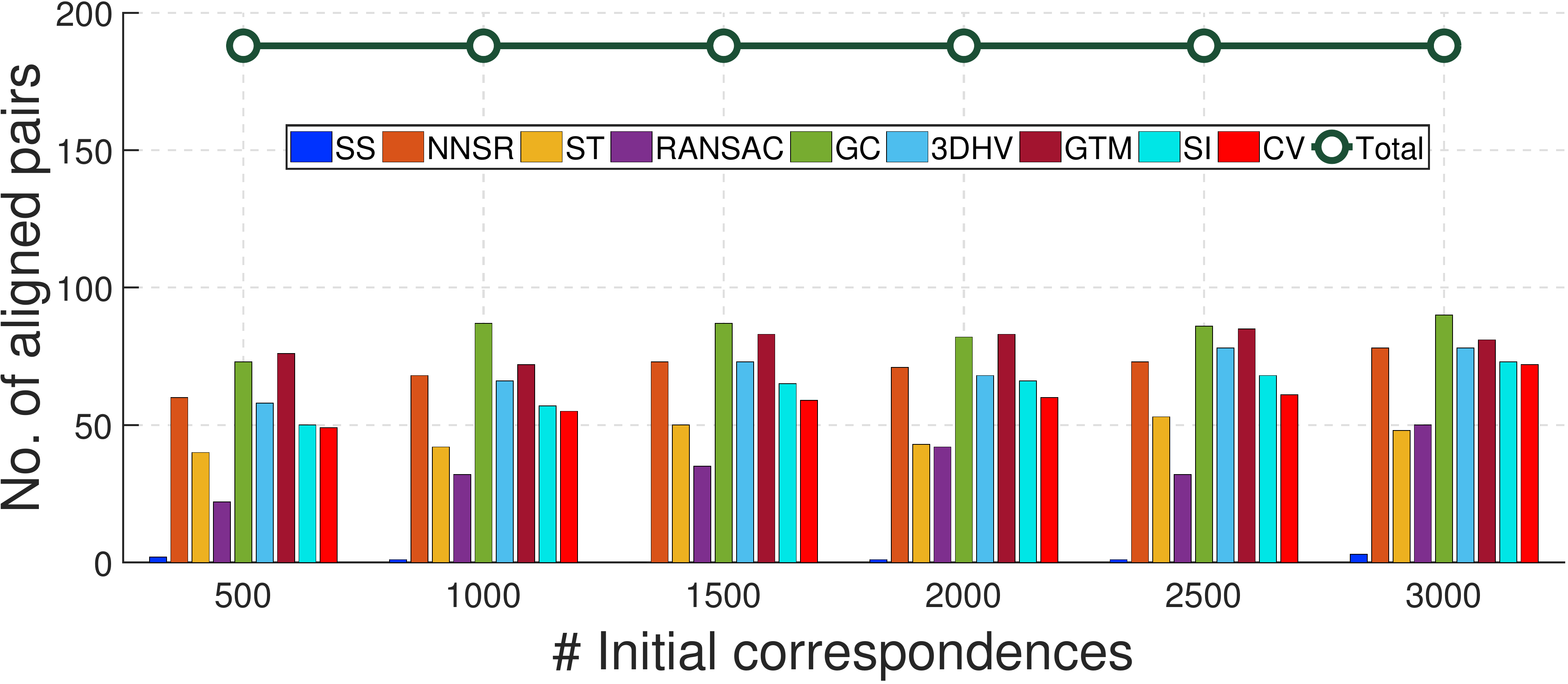}}
	\end{minipage}
	\begin{minipage}{0.47\linewidth}
		\raggedright
		\subfigure[\textit{Det.} \&  \textit{Desc. on B3R }]{
			\includegraphics[width=0.85\linewidth]{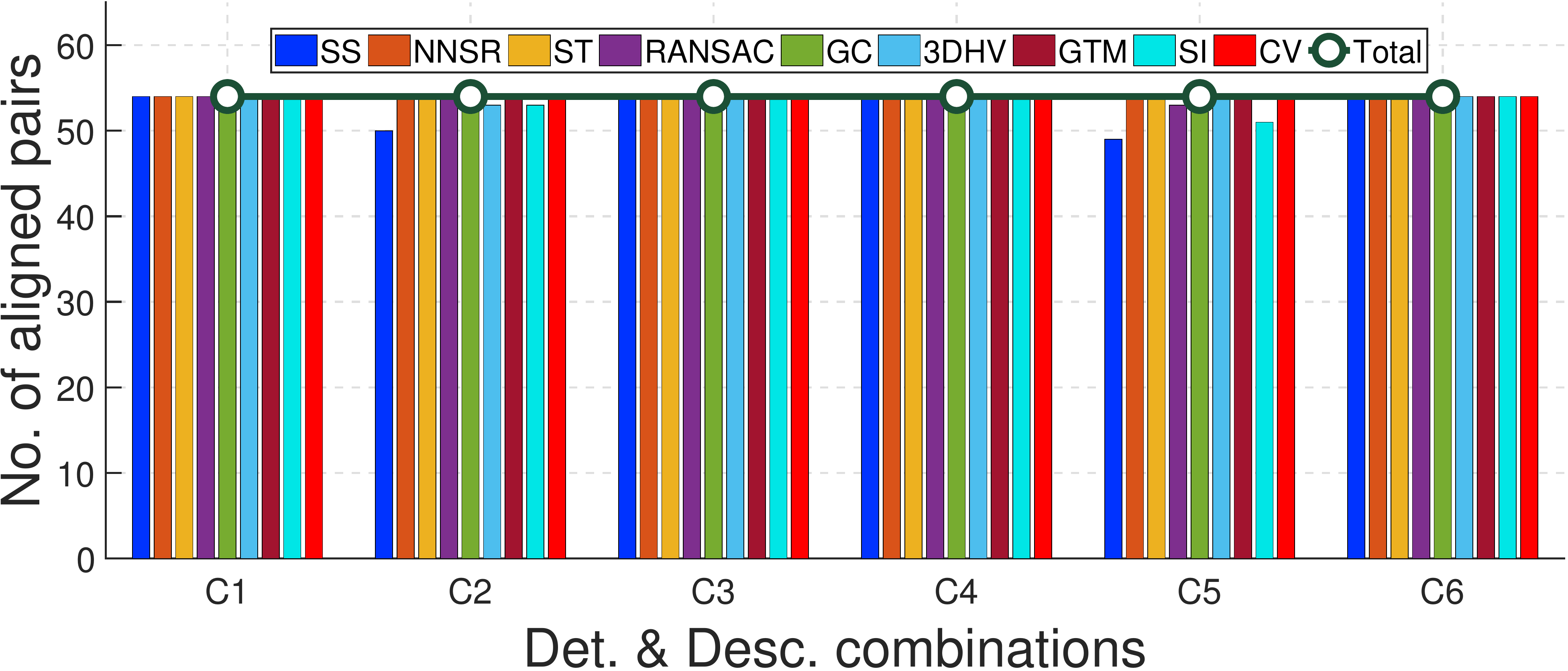}}
	\end{minipage}
	\begin{minipage}{0.47\linewidth}
		\raggedleft
		\subfigure[\textit{Det.} \&  \textit{Desc. on U3OR }]{
			\includegraphics[width=0.85\linewidth]{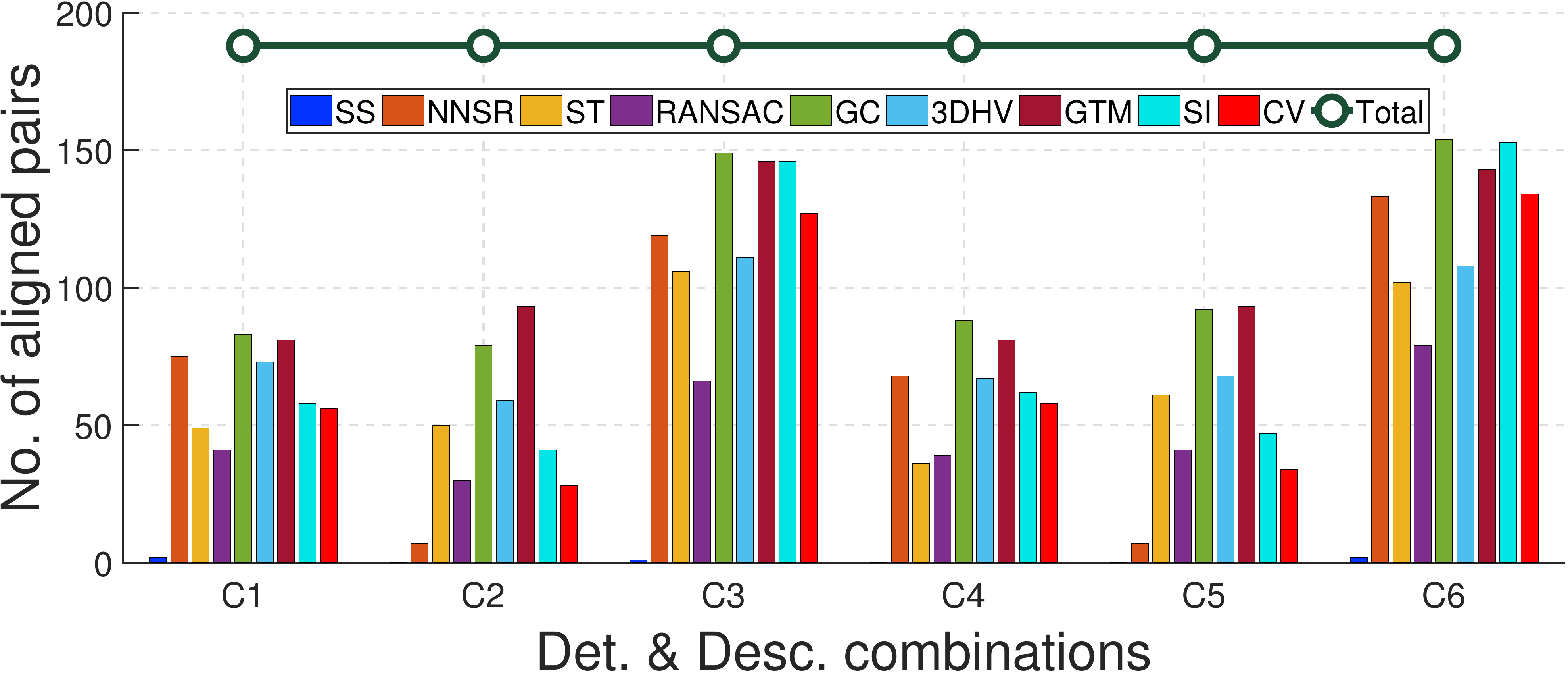}}
	\end{minipage}
	\hfill
	\begin{minipage}{0.47\linewidth}
		\raggedright
		\subfigure[\textit{Det.} \&  \textit{Desc. on U3M }]{
			\includegraphics[width=0.85\linewidth]{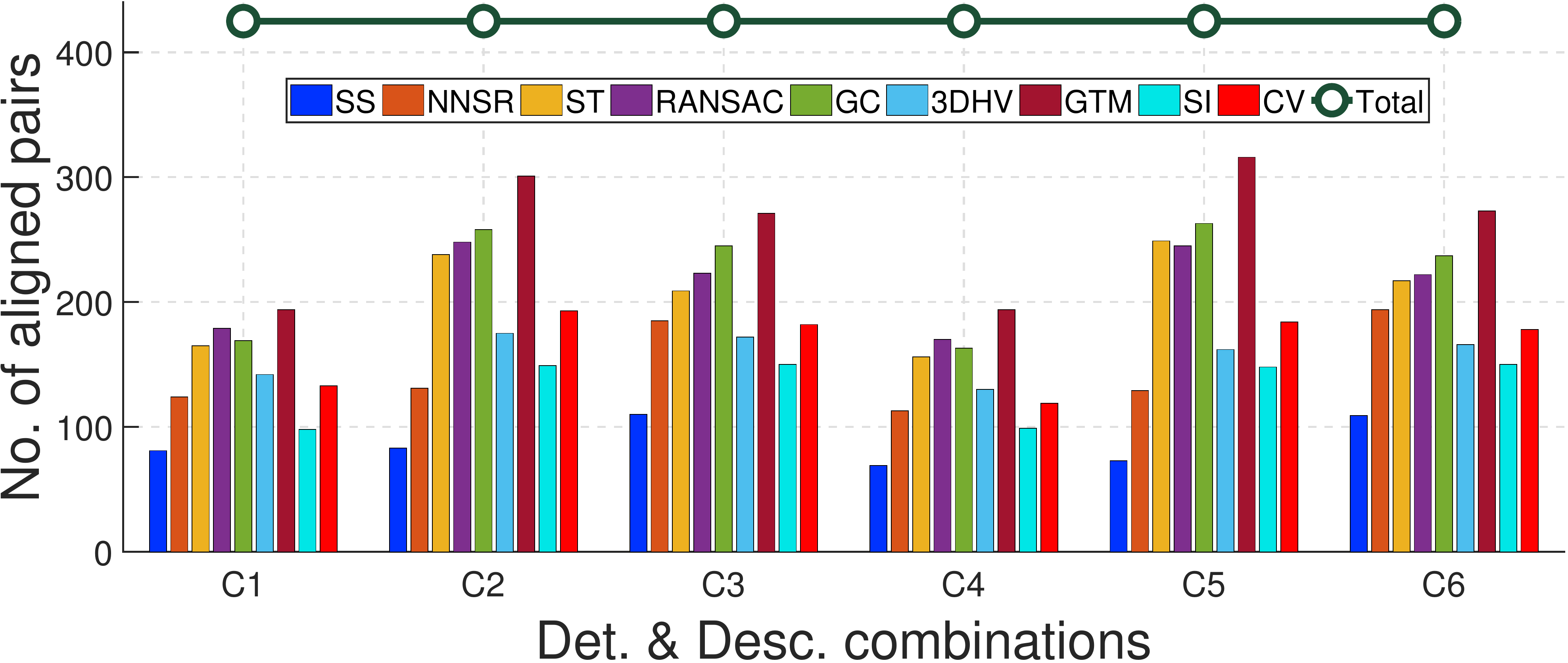}}
	\end{minipage}
	\hfill
	\caption{Rigid registration performance of selected methods in Sect.~\ref{sec:eval_mtd} regarding various application scenarios and challenges (Sect.~\ref{sec:challenges}).}
	\label{fig:reg}
\end{figure*}
\begin{enumerate}[]
\item  Fig.~\ref{fig:PRF}(a) shows that when the standard deviation of noise is smaller than 0.2 \textit{pr}, all methods achieve acceptable performance where RANSAC and ST are two top-ranked ones. The superiority of these two methods becomes more obvious as the noise gets severe. In terms of precision, it is interesting to note that NNSR even surpasses GC and 3DHV. It is because NNSR prefers to select distinctive correspondences,
which is peculiar sufficient in this dataset as	the models possess wealthy distinctive structures. ST, RANSAC, GC, GTM, and CV manage to retrieve almost all inliers under all levels of noise. A significant deterioration of performance can be found for SI when undergoing severe noise, indicating its sensitivity to such perturbation.

\item   Fig.~\ref{fig:PRF}(b) shows some similar results to Fig.~\ref{fig:PRF}(a). For instance, RANSAC and ST are still the  best two competitors and a high recall is achieved by many methods such as GTM and CV. Yet, the differences are that SS even outperforms SI when the down-sampling ratio reaches 0.3 regarding precision and the recall performance	of NNSR and SI drops dramatically for sparse data. It arises from the SHOT's sensitivity to varying point densities~\cite{tombari2010unique}, making the feature less distinctive (e.g., NNSR's principle) for data with significant resolution variation. The reason for SI is that the LRF of SHOT (e.g., the component in the global voting stage
for SI) exhibits poor repeatability with data resolution variation~\cite{yang2018toward}.

\item Fig.~\ref{fig:PRF}(c) and Fig.~\ref{fig:PRF}(d) suggest that all methods meet a dramatic performance degradation. By referring to Fig.~\ref{fig:inlier_info} we can find that U3OR dataset is more challenging than B3R dataset. A salient point can be found that GTM is the top performer, which exceeds the second best one by a large gap  in terms of precision under the condition of clutter and occlusion. It is due to the fact that GTM has very strong selectivity as it adopts $L_1$-type constraint to optimize the inlier searching problem~\cite{rodola2013elastic}. Since the recall performance of GTM is mediocre, we can conclude that the reduced correspondence set by GTM is sparse but accurate. Regarding the ranking of recall performance, CV is the best method and its gap with others is more clear in cluttered scenes. Notably, the ST method, with leading performance on the B3R dataset, performs
quite poor on the U3OR dataset. This is because ST tries to	find large isometry-maintained clusters, which rarely exit	in scenes with high percentages of clutter and occlusion.

\item Common to all algorithms, as shown in Fig.~\ref{fig:PRF}(e),  is that their performance generally drops as the degree of overlap decreases. When the percentage of overlap is greater than 70\%, GTM significantly outperforms other in terms of precision. More specifically, it surpasses the second best one by approximately 30\% and 20\% with 90\% and 80\% overlap, respectively. As the overlap ratio decreases, GTM and RANSAC achieve comparable precision performance. In terms of recall, CV outperforms all others at all levels of overlap, followed by SI. A common trait for CV and SI is that both methods are individual-based and they employ a voting mechanism to judge the correctness of a correspondence.  So they have no requirements on the spatial distribution of inliers and those isolated inliers can be also identified by these two methods, giving a rise in recall.  Weighing up both precision and recall, GTM and RANSAC are more preferable than others.

\item When changing the judging threshold for an inlier, as shown in Fig.~\ref{fig:PRF}(f), all methods achieve a performance gain in terms of precision, where RANSAC and GTM climb more rapidly. However, only SI, CV, and SS achieve an evident performance improvement regarding recall. The ranking of all methods is generally stable as the judging threshold changes.

\item Fig.~\ref{fig:PRF}(g) suggests that different algorithms give different responses when varying the number of initial feature matches. The performance of some algorithms, e.g., GC, RANSAC, and 3DHV, fluctuates as the number of initial matches augments. Meanwhile, one can find that the size of initial correspondence set has a relatively strong impact
on the SI and CV algorithms. To be more specific, when there are less than 1000 input correspondences, these two algorithms suffer from low precision values. However, as
the initial feature matches become dense, i.e., more than 1000 correspondences, the precision performance of SI and CV improves quickly. Note that the CV method even reaches
the second best precision with around 3000 initial correspondences. Still, SI and CV are the best two methods in terms of recall performance and surpass the others by a clear gap.

\item The effect of changing the combination of keypoint detector and descriptor is reflected by Fig.~\ref{fig:PRF}(h)-(j). On the B3R dataset, using H3D+LFSH and ISS+LFSH for computing initial correspondences results in far less inliers than others, but GTM still achieves near 100\% precision and exceeds the others under all considered combinations of detector and descriptor. CV, GC, and RANSAC manage to retrieve almost all inliers within the initial set regardless the change of ``detector-descriptor'' combination. On the U3OR dataset, the best performance is achieved by all methods when using RCS for local geometric feature description. The ranking of tested methods is generally consistent when varying detectors and descriptors. Specifically, GTM achieves the best precision performance due to its outstanding selectivity and both SI and CV obtain superior recall performance. On the U3M dataset, the best performance of most methods is achieved when using ISS+LFSH.  The ranking of SI, 3DHV, and NNSR sometimes alters with the change of detector and descriptor but the ranking of other methods is quite consistent. In particular, GTM achieves the best overall performance, followed by GC, RANSAC, and ST.
\end{enumerate}

To summarize, RANSAC and ST are viable choices in the presence of noise and data density variation. In terms of robustness to clutter, occlusion, and partial overlap, GTM is able to find a reduced set with particularly high precision, though its recall performance is inferior than most of the other methods. SI and CV, as two individual-based methods with a voting scheme for correspondence selection, are able to retrieve the majority of inliers under all tested conditions.
\subsection{Rigid Registration Performance}
The rigid registration performance of evaluated methods according to the protocol in Sect.~\ref{sec:metric} is shown in Fig.~\ref{fig:per_p}. Several observations can be made from the figure.
\begin{enumerate}[]
\item All methods manage to align the data pairs in the B3R dataset with Gaussian noise ( Fig.~\ref{fig:per_p}(a)). This can be explained by Fig.~\ref{fig:PRF}(a) because all methods obtain an average precision value that is greater than 0.4. When undergoing data decimation ( Fig.~\ref{fig:per_p}(b)), all methods achieve a 100\% registration precision when the down-sampling ratio is greater than 0.3, but some methods (e.g., SI and SS) start to show failure cases as the data are further simplified. Most methods achieve pleasurable registration performance on the B3R dataset with noise and density variation. 
\item In the presence of clutter and occlusion (Fig.~\ref{fig:per_p}(c)-(d)), GC achieves the best performance, followed by GTM. This deviates much with the rankings presented in Fig.~\ref{fig:PRF}(c)-(d). For example, RANSAC gets higher precision than GC in most cases but obtains inferior registration performance. Similar observations can be found for GTM and NNSR. The reason is that some group-based methods (e.g., GTM and RANSAC) assume that inliers exist in a cluster form and the precision of the retrieved set by these methods is often extremely low or high. More intuitively, as shown in Fig.~\ref{fig:per_p}, GC rarely achieves a precision value greater than 0.5 per matching pair on the U3OR dataset, but the correspondence set selected by GC always contains more than 10\% inliers. By contrast, RANSAC achieves particularly high precisions (usually greater than 0.5) for successful instances but may not able to retrieve any inliers in failure cases. This leads to the result that GC behaves better than RANSAC in terms of rigid registration performance.
\begin{figure}[t]
	\centering
	\includegraphics[width=\linewidth]{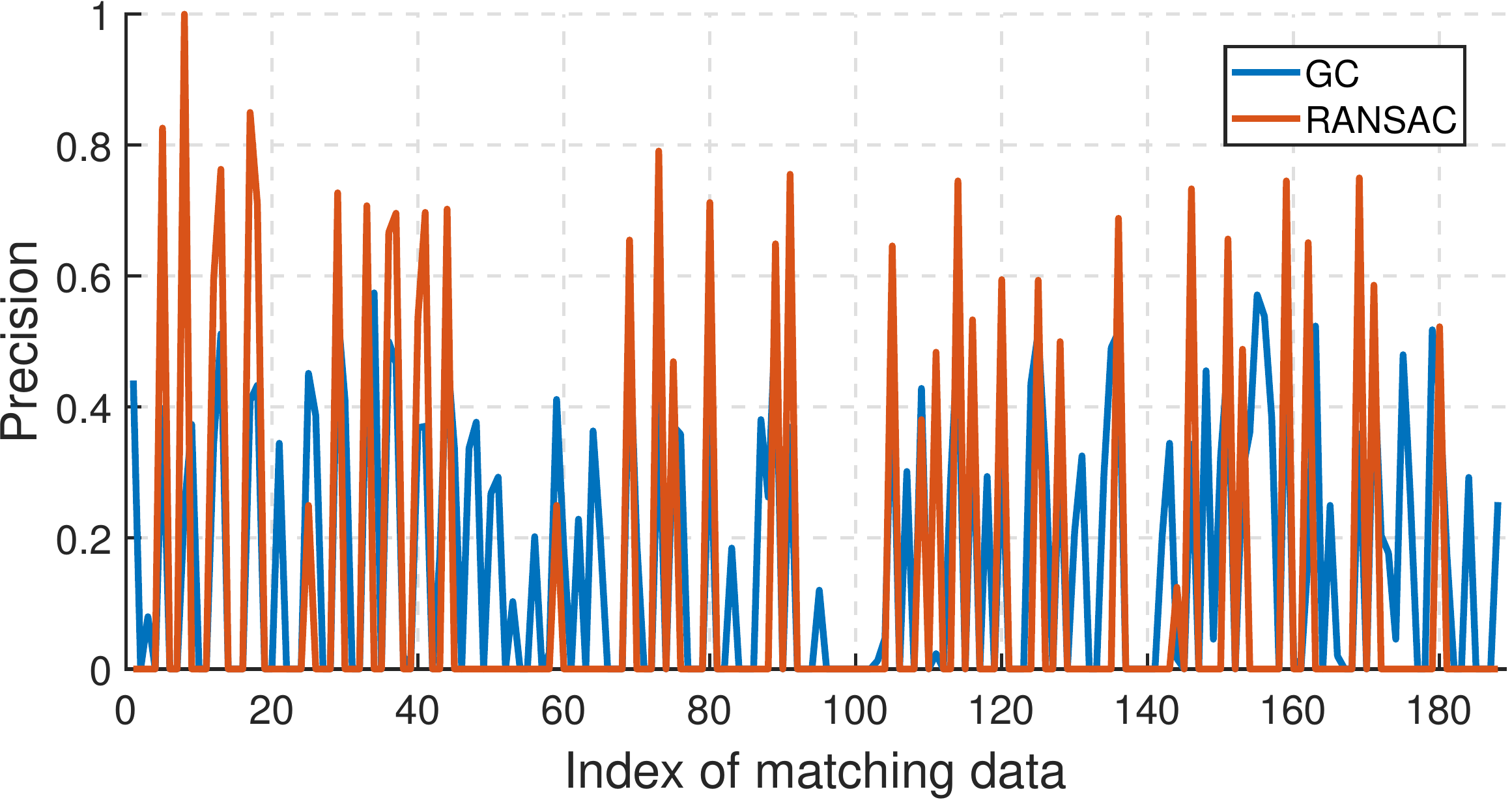}\\
	\caption{Precision results of GC and RANSAC on each matching pair from the U3OR dataset.}
	\label{fig:per_p}
\end{figure}
\item When aligning data pairs with partial overlap (Fig.~\ref{fig:per_p}(e)), GTM achieves the best performance under all levels of partial overlap, followed by RANSAC and GC. When the overlap ratio is smaller than 0.6, all methods fail to align most of the data pairs. As will later be shown in Fig.~\ref{fig:per_p}(j), the overall registration performance is closely related to the selection of a proper ``detector-descriptor'' combination. 
\item The change of threshold $\epsilon$ (Fig.~\ref{fig:per_p}(f)) has a  faint effect on the overall ranking of evaluated methods respecting registration performance, where GC and GTM are the best two methods. When changing the size of the initial correspondence set, as shown in Fig.~\ref{fig:per_p}(g), the increase of the initial correspondence count slightly boosts the rigid registration performance of most methods. GTM is the best competitor with around 500 initial matches and is surpassed by GC with denser matches.
\item For considered ``detector-descriptor'' combinations, all methods achieves stable yet high registration precision on the B3R dataset (Fig.~\ref{fig:per_p}(h)). On the U3OR dataset (Fig.~\ref{fig:per_p}(i)), GC and GTM generally deliver the best performance. Notably, SS shows significantly poor performance for all combinations. This is because most models in the U3OR dataset are symmetric, resulting in many repetitive patterns that are hard to distinguish simply using the feature similarity cue. Nonetheless, using feature similarity ratio, i.e., NNSR, sometimes presents very competitive performance.   Fig.~\ref{fig:per_p}(j) shows that GTM outperforms the others in all cases and the peak is achieved with correspondences generated by ISS+LFSH. 
\end{enumerate}

In general, rigid data in the shape retrieval context even with noise and density variation can be accurately aligned using most of the evaluated methods. For scenarios such as 3D object recognition and point cloud registration, GC and GTM present more competitive performance than the others.

\subsection{Computational Efficiency}
\begin{figure}[t]
	\centering
	\includegraphics[width=\linewidth]{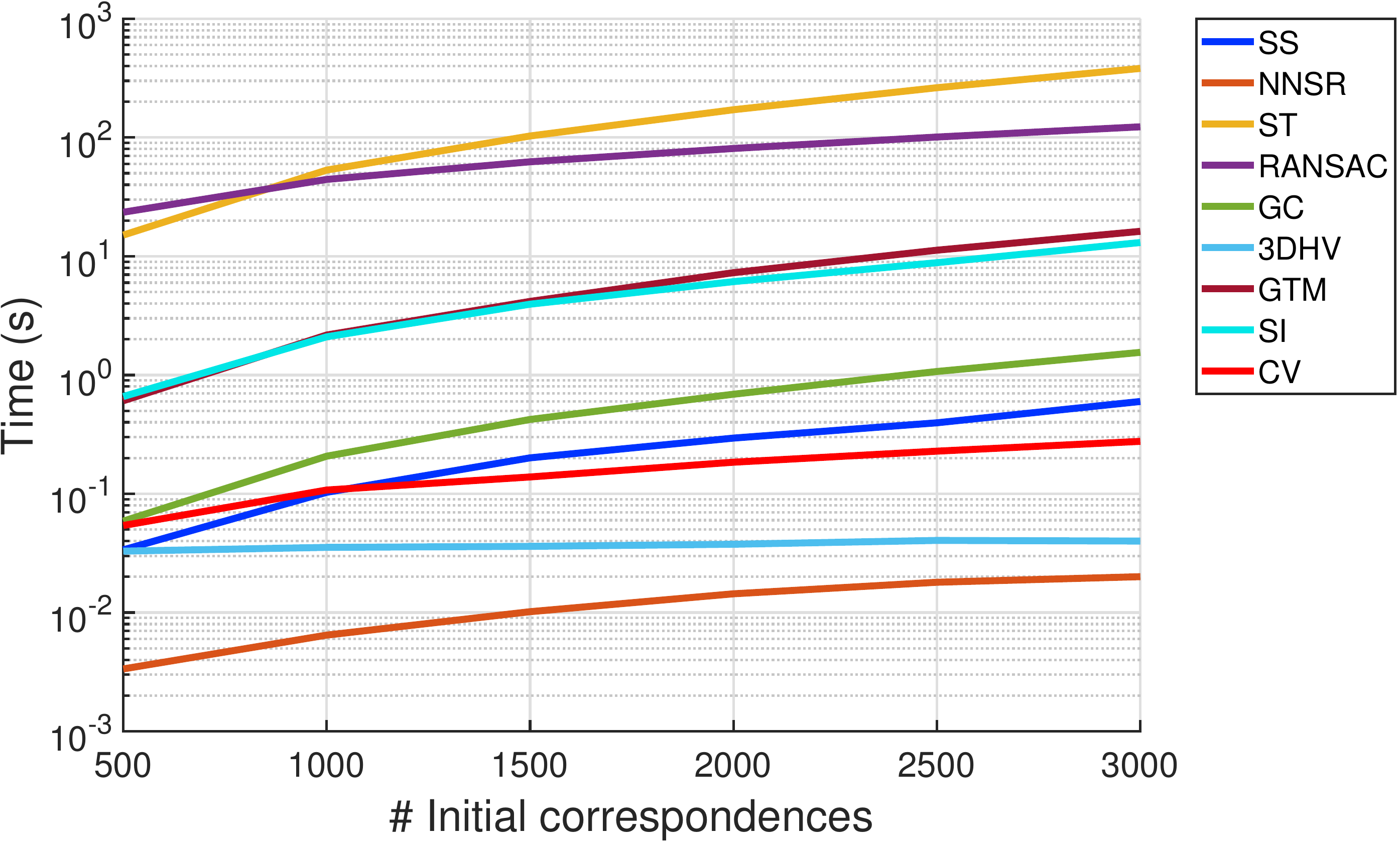}\\
	\caption{Computational efficiency with regards to different  sizes of the initial correspondence set.}
	\label{fig:time}
\end{figure}

Respecting the efficiency concern of a correspondence grouping method, the time cost is mainly affected by the number of initial correspondences that often varies with application contexts and the selection of keypoint detector. Therefore it is necessary to compare the computational efficiency of considered methods with respect to inputs with different cardinalities. The deployment of this experiment is as follows. First, the NMS radius of the Harris 3D keypoint detector is varied to obtain initial correspondence sets with different scales. Second, these sets are fed to the evaluated methods and their time costs are recorded. Finally, we repeat the former stage 10 times  and record the average timing results. The efficiency evaluation result is shown in Fig.~\ref{fig:time}. We also provide the computational complexity information of evaluated methods in Table~\ref{tab:time_complex} to help explaining the result. 

\begin{table}[t]]\small
	\renewcommand{\arraystretch}{1}
	\caption{Computational complexity of evaluated methods.}
	\label{tab:time_complex}
	\centering
	\begin{tabular}{cc}
		\hline
		SS  & $O(n{\rm log}n)$\\
		\hline
		NNSR  & $O(n)$\\
		\hline
		ST  & $O(n^3)$\\
		\hline
		RANSAC  &$O(n^3)$\\
		\hline
		GC  & $O(n^2)$\\
		\hline
		3DHV  & $O(n)$\\
		\hline
		GTM  & $O(n^2)$\\
		\hline
		SI  & $O(n{\rm log}n)$\\
		\hline
		CV&$O(n{\rm log}n)$\\
		\hline
	\end{tabular} 
\end{table}
One can make several observations from the results. First, ST and RANSAC are the two most time-consuming ones, especially for large-scale initial correspondence sets. The reason for ST is that the computational cost for solving the principle eigenvector of an $n \times n$ matrix increases dramatically when the order $n$ (i.e., the size of the initial correspondence set) gets larger. The explanation for RANSAC is that ``hypothesis-verification'' is repeatedly performed to search the inlier cluster and at each iteration the whole correspondence set is required to participate the calculation of retrieved inliers. Second, NNSR and 3DHV are the two most efficient ones. We remark that as SS employs an adaptive thresholding strategy~\cite{otsu1975threshold} in our implementation, it is therefore a bit less efficient than NNSR. 3DHV is efficient even for correspondence sets with thousands of candidates due to the linear computational complexity. Third, GTM and SI are more time-consuming than most compared methods. It is because GTM has a computational complexity of $O(n^2)$~\cite{rodola2013scale} and SI needs local and global consolidations to judge the correctness of a correspondence. Although CV  is a voting-based method as well, far less time is required because CV does not conduct nearest neighbor search to find voters, as opposed to the local voting stage in SI.
\subsection{Visualization}
\begin{figure*}[t]
	\centering
	\includegraphics[width=0.95\linewidth]{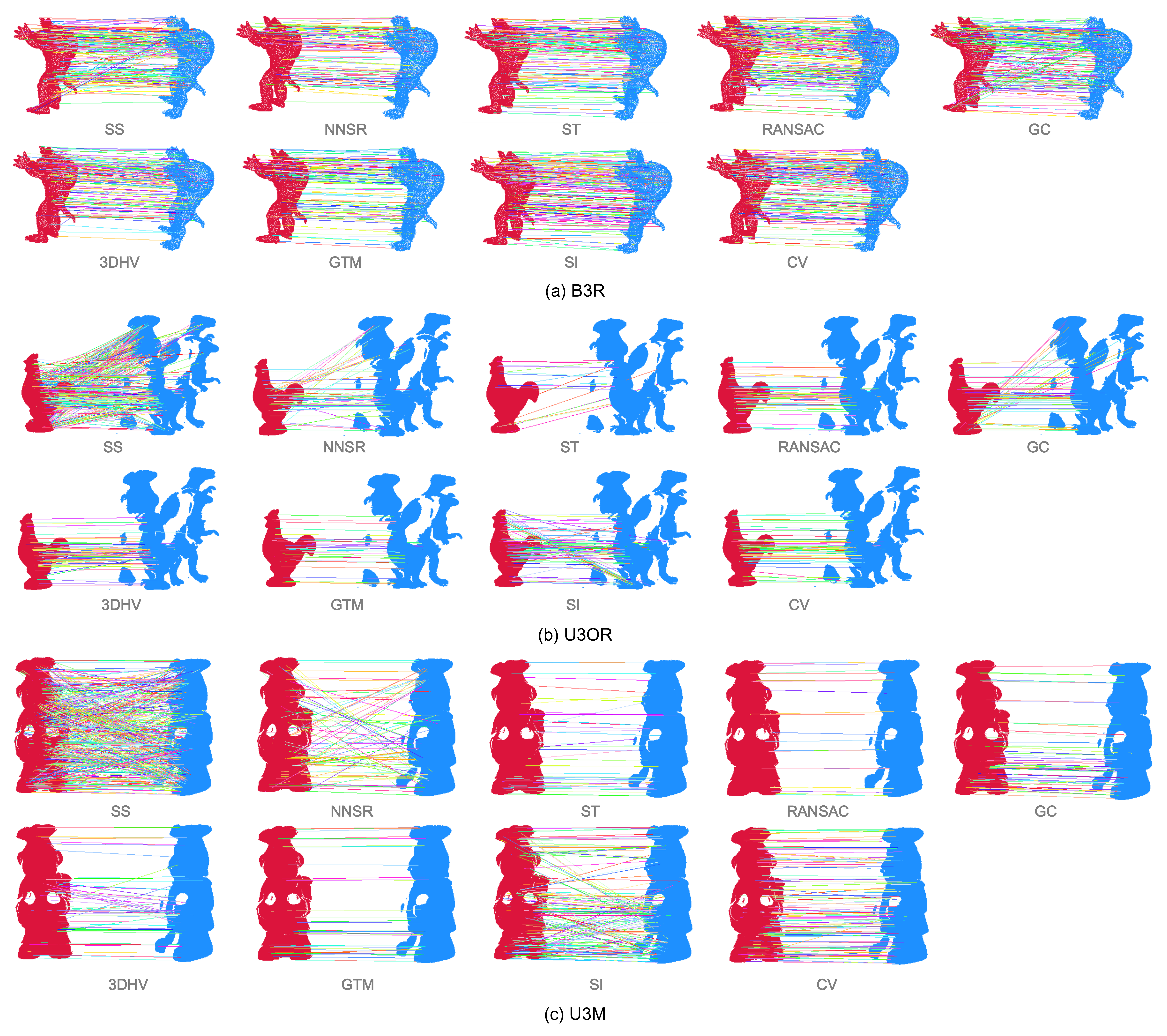}\\
	\caption{Visual results of the grouped inlier sets by evaluated methods on sample data from the experimental datasets.}
	\label{fig:visual}
\end{figure*}
We finally provide some visual results of the selected correspondence sets by evaluated methods in Fig.~\ref{fig:visual}. From the figure, we can percept some visual differences of  outcomes by different methods. For instance, the selected sets by SS and NNSR contain a relatively high ratio of outliers except  on the B3R dataset. This verifies that algorithms relying on feature matching score are very sensitive to nuisances directly affecting a feature's discriminative power, e.g., clutter, occlusion, and partial overlap.  Another evident phenomenon is that with different grouping principles, the number as well as the spatial locations of the outcomes of evaluated methods generally differ from each other.
\section{Summary and Discussion}\label{sec:sum}
In light of the evaluation results shown in Sect.~\ref{sec:result},  we present a summary of evaluated methods regarding their advantages, shortcomings, and suitable applications in the following.

\begin{itemize}
	\item \textbf{SS} and \textbf{NNSR}, as two baselines relying on the distinctiveness of the local geometric descriptor, are sensitive to perturbations such as clutter, occlusion, and partial overlap. For data with rich geometric information, NNSR can be an intriguing option as it also affords real-time performance.
	\item \textbf{ST} has a main positive trait that both high precision and recall can be achieved when sufficient inliers are included in the raw correspondence set. However, its performance degrades significantly in 3D object recognition and point cloud registration applications that usually suffer from particularly low inlier ratios. ST is also  not competitive for time-crucial applications because it is the most time-consuming one among evaluated methods especially for large-scale correspondence sets.
	\item \textbf{RANSAC} behaves well with high quality initial matches, but often achieves extremely low precisions for inputs with scarce inliers. This makes it suffer from relatively poor rigid registration performance in 3D object recognition context. Similar to ST, performing correspondence selection with RANSAC on large-scale dense correspondence set is inefficient.
	\item \textbf{GC} is recommended for rigid data registration problems since it shows superior registration performance in various application scenarios. For correspondence grouping, GC is inferior to most methods when the initial inlier ratio is high but delivers pleasurable performance for correspondence sets with low inlier ratios.
	\item \textbf{3DHV} is an ultra efficient algorithm which simultaneously returns acceptable inlier searching performance	in many applications. These merits suggest that 3DHV can be applied to time-crucial applications, e.g., simultaneous
	localization and mapping (SLAM), object	grasping, and 3D object recognition in mobile platforms.
	\item \textbf{GTM} is an outstanding method with superior precision results in terms of correspondence grouping and rigid data registration. The gap with other methods is even more evident in challenging conditions such as object recognition with clutter and occlusion and data registration with limited overlap. The shortcomings of GTM include the limited recall performance and expensive timing cost when coping with dense initial correspondences.
	\item \textbf{SI} and \textbf{CV} are preferable choices for applications requiring dense correspondences because both methods are prominent in terms of recall under various conditions. Specifically, CV shows better precision performance and is more computationally efficient than SI.
\end{itemize}
When comparing \textit{group}-based and \textit{individual}-based methods from a general perspective, we can summarize that more promising precision performance can be achieved by group-based methods and individual-based methods (except for SS and NNSR that merely leverage the local geometric feature cue) exhibit better recall performance. Our evaluation also confirms the necessity of performing correspondence grouping because directly performing RANSAC on the raw correspondence set results in poor registration performance, while using proper correspondence selection methods such as GC and GTM prior to transformation estimation can improve the performance greatly. Finally, two open issues are highlighted by this evaluation.
\begin{itemize}
		\item More effective solutions to correspondence selection in 3D object recognition and point cloud registration contexts. Fig.~\ref{fig:PRF} and Fig.~\ref{fig:reg} consistently suggest that all tested methods deliver limited performance with severe clutter and occlusion or very limited overlapping region. This is because of the resulted extremely low inlier ratio (Fig.~\ref{fig:inlier_info}), which remains a great challenge in this research area.
		\item A good trade-off among precision, recall, and efficiency. All evaluated methods in this paper cannot balance well regarding the three aspects, which may limit their deployments in real-world applications.
\end{itemize}
\section{Conclusions}\label{sec:conc}
This paper has presented a comprehensive performance evaluation of several state-of-the-art 3D correspondence selection methods. We have also abstracted all considered methods to a number of core stages to better interpret their peculiarities and differences. The experiments were deployed on datasets with a variety of application scenarios and nuisances, thus leading to raw correspondence sets for grouping with different numbers of inliers, inlier ratios, and spatial locations. Furthermore, we have summarized the traits, merits, and demerits of each method based on the evaluation results. Our evaluation therefore presents valuable guidances to developers for the choice of a proper correspondence selection method regarding a particular application, and may help the following scholars to highlight the blind spots in this area and devise effective methods to break current bottlenecks.
\section*{Acknowledgments}
The authors would like thank Dr. Anders Glent Buch and Dr. Emanuele Rodol\`{a} for sharing the code of their methods to us, and the developers of PCL for making many methods evaluated in this paper publicly available. We would also like to thank the Standford 3D Scanning Repository, the University of Western Australia, and the University of Bologna for freely providing their datasets. This work is supported in part by the National Natural Science Foundation of China (No. 61876152).
\bibliographystyle{IEEEtran}
\bibliography{mybibfile}



\end{document}